\documentclass{article}
\usepackage{authblk}
\usepackage{amsmath, amssymb}
\usepackage[margin=1in]{geometry}
\usepackage[subrefformat=parens,labelformat=parens]{subfig}

\usepackage{todonotes}
\usepackage{appendix}
\usepackage{lineno}
\usepackage{url}
\usepackage{xcolor}
\usepackage{caption}

\newcommand{\Tr}{\text{Tr}}
\title{Design of optical neural networks with component imprecisions}
\author[1, 2, 3]{Michael Y.-S. Fang \thanks{michaelfang@berkeley.edu}}
\author[4]{Sasikanth Manipatruni}
\author[5]{Casimir Wierzynski}
\author[5]{Amir Khosrowshahi}
\author[1, 2, 6]{Michael R. DeWeese}
\affil[1]{Department of Physics, University of California Berkeley, Berkeley, CA, USA}
\affil[2]{Redwood Center for Theoretical Neuroscience, University of California Berkeley, Berkeley, CA, USA}
\affil[3]{Molecular Biophysics \& Integrated Bioimaging, Lawrence Berkeley National Lab, Berkeley, CA, USA}
\affil[4]{Components Research, Intel Corporation, Hillsboro, OR, USA}
\affil[5]{Intel AI Research, San Diego, CA, USA}
\affil[6]{Helen Wills Neuroscience Institute, University of California Berkeley, Berkeley, CA, USA}
\date{}
 
\begin{document}
\maketitle



\begin{abstract}
For the benefit of designing scalable, fault resistant optical neural networks (ONNs), we investigate the effects architectural designs have on the ONNs' robustness to imprecise components. We train two ONNs -- one with a more tunable design (GridNet) and one with better fault tolerance (FFTNet) -- to classify handwritten digits. When simulated without any imperfections, GridNet yields a better accuracy ($\sim 98 \%$) than FFTNet ($\sim 95 \%$). However, under a small amount of error in their photonic components, the more fault tolerant FFTNet overtakes GridNet. We further provide thorough quantitative and qualitative analyses of ONNs' sensitivity to varying levels and types of imprecisions. Our results offer guidelines for the principled design of fault-tolerant ONNs as well as a foundation for further research.
\end{abstract}

%%%%%%%%%%%%%%%%%%%%%%%%%% body %%%%%%%%%%%%%%%%%%%%%%%%%%
\section{Introduction}
\label{sec:intro}
Motivated by the increasing capability of artificial neural networks in solving a large class of problems, optical neural networks (ONNs) have been suggested as a low power, low latency alternative to digitally implemented neural networks. A diverse set of designs have been proposed, including Hopfield networks with LED arrays \cite{farhat1985optical}, optoelectronic implementation of reservoir computing\cite{paquot2012optoelectronic, appeltant2011information}, spiking recurrent networks with microring resonators\cite{tait2017neuromorphic, tait2014broadcast}, convolutional networks through diffractive optics\cite{chang2018hybrid}, and fully connected, feedforward networks using Mach-Zehnder interferometers (MZIs) \cite{shen2017deep}. 

We will focus on the last class of neural networks, which consist of alternating layers of modules performing linear operations  and element-wise nonlinearities\cite{goodfellow2016deep}. The $N$-dimensional complex-valued inputs to this network are represented as coherent optical signals on $N$ single-mode waveguides. Recent research into configurable linear optical networks\cite{reck1994experimental, clements2016optimal, barak2007quantum, carolan2015universal, harris2017quantum} enables the efficient implementation of linear operations with photonic devices. These linear multipliers, layered with optical nonlinearities form the basis of the physical design of ONNs. In Sec. \ref{sec:physical}, we provide a detailed  description of two specific architectures -- GridNet and FFTNet -- both built from MZIs.

While linear operations are made much more efficient with ONNs in both power and speed, a major challenge to the utility of ONNs lies in their susceptibility to fabrication errors and other types of imprecisions in their photonic components. Therefore, realistic considerations of ONNs require that these imprecisions be taken into account. Previous analyses of the effects of fabrication errors on photonic networks were in the context of post-fabrication optimization of unitary networks \cite{pai2018matrix,russell2017direct,burgwal2017using}. Our study differs in three main areas. 

First, In the previous work, unitary optical networks were optimized to simulate randomly sampled unitary matrices. We, instead, train optical neural networks to classify structured data. ONNs, in addition to unitary optical multipliers, include nonlinearities, which add to its complexity. 

Second, rather than optimization towards a specific matrix, the linear operations learned for the classification task is not, \emph{a priori}, known. As such, our primary figure of merit is the classification accuracy instead of the fidelity between the target unitary matrix and the one learned.

Lastly, the aforementioned studies mainly focused on the optimization of the networks after fabrication. The imprecisions introduced generally reduced the expressivity of the network -- how well the network can represent arbitrary transformations. Evaluation of this reduction in tunability and mitigating strategies were provided. However, such post-fabrication optimization requires the characterization of every MZI, the number of which scales with the dimension ($N$) of the network as $N^2$. Protocols for self configuration of imprecise photonic networks have been demonstrated \cite{miller2015perfect, wilkes201660}. While measurement of MZIs were not necessary in such protocols, each MZI needed to be configured progressively and sequentially. Thus, the same $N^2$ scaling problem remained. Furthermore, if multiple ONN devices are fabricated, each device, with unique imperfections, has to be optimized separately. The total computational power required, therefore, scales with the number of devices produced.

In contrast, we consider the effects of imprecisions introduced after software training of ONNs (Code 1, Ref. \cite{Fang2019Code}), details of which we present in Sec. \ref{sec:software}. This pre-fabrication training is more scalable, both in network size and fabrication volume. An ideal ONN (i.e., one with no imprecisions) is trained in software only once and the parameters are transferred to multiple fabricated instances of the network with imprecise components. No subsequent characterization or tuning of devices are necessary. In addition to the benefit of better scalability, fabrication of static MZIs can be made more precise and cost effective compared to re-configurable ones.

We evaluate the degradation of ONNs from their ideal performances with increasing imprecision. To understand how such effects can be minimized, we investigate the role that the architectural designs have on ONNs' sensitivity to imprecisions. The results are presented in Sec. \ref{sec:accuracy}.  Specifically, we study the performance of two ONNs in handwritten digit classification. GridNet and FFTNet are compared in their robustness to imprecisions. We found that GridNet achieved a higher accuracy ($\sim 98\%$) when simulated with ideal components compared to FFTNet ($\sim 95\%$). However, FFTNet is much more robust to imprecisions. After the introduction of realistic levels of error, the performance of GridNet quickly degrades to below that of FFTNet. We also show, in detail, the effect that specific levels of noise has on both networks. 

In Sec. \ref{sec:stack_trunc}, we demonstrate that this is due to more than the shallow depth of FFTNet and that FFT-like architectures is more robust to error when compared to Grid-like architectures of the same depth.
%One main reason that FFTNet is more robust than GridNet is because of its much shallower depth. However, in Sec. \ref{sec:stack_trunc}, we demonstrate that the robustness of FFTNet is due to more than just its shallow depth.

In Sec. \ref{sec:pos_sen}, we investigate the effects localized imprecisions have on the network by constraining the imprecisions to specific groups of MZIs. We demonstrate that the network's sensitivity to imprecisions is dependent on algorithmic choices as well as its physical architecture. 

With a growing interest in optical neural networks, a thorough analysis of the relationship between ONNs' architecture and its robustness to imprecisions and errors is necessary. From the results that follow, in this article, we hope to provide a reference and foundation for the informed design of scalable, error resistant ONNs.

\section{Physical design of optical neural networks}
\label{sec:physical}
\begin{figure}[h!]
    \centering
    \subfloat[GridNet linear layer]{%0.1
    \label{fig:grid_arch}%0.1
    \includegraphics[height=1.5in]{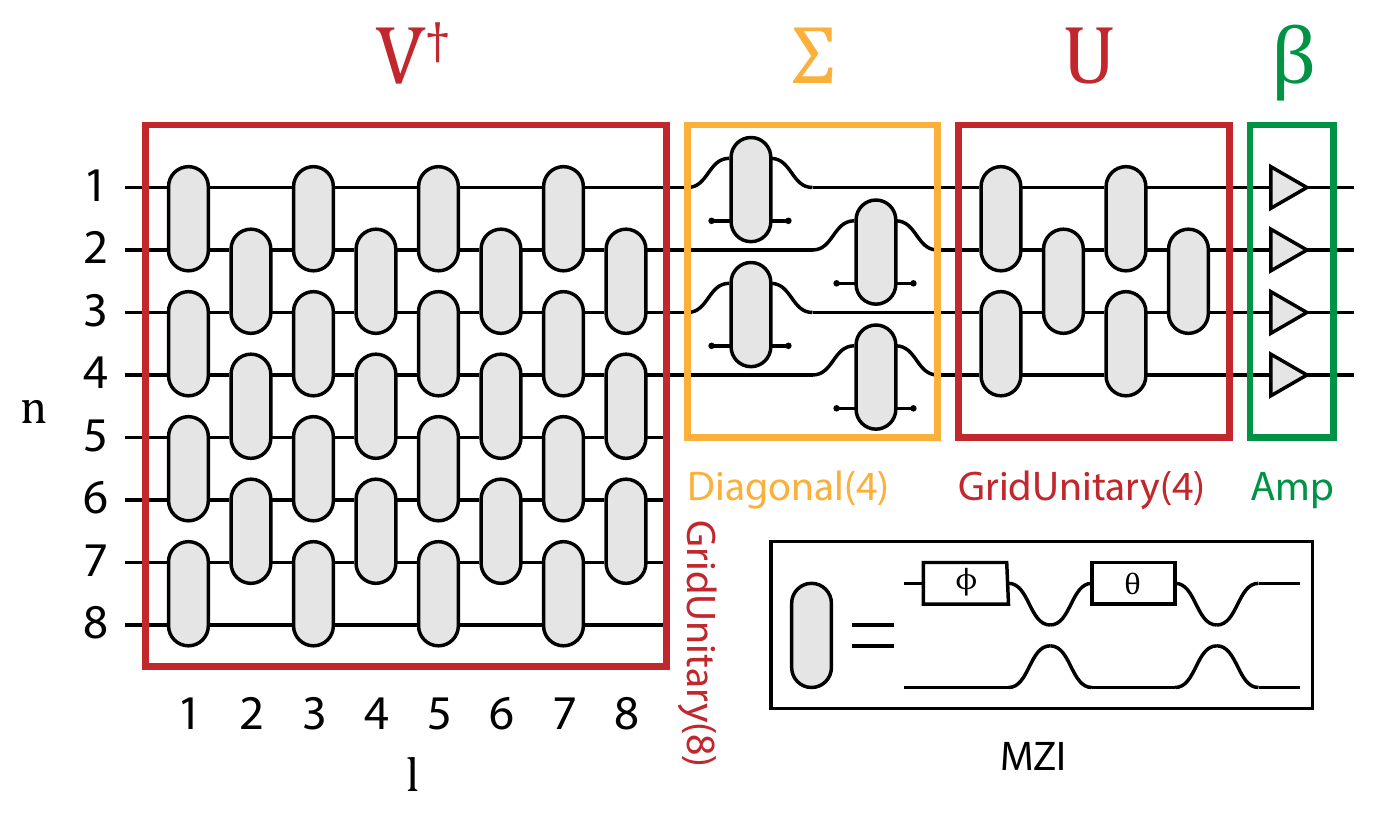}}%
    \subfloat[FFTNet linear layer]{%0.81
    \label{fig:fft_arch}%0.81
    \includegraphics[height=1.5in]{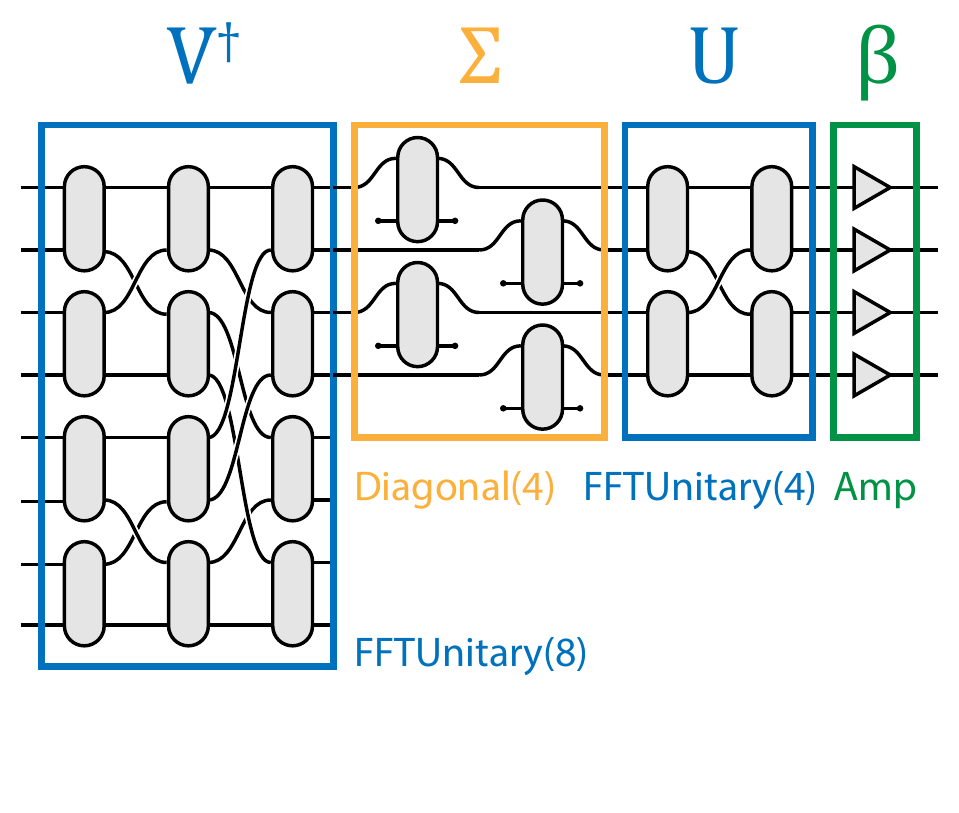}}%
    \caption{a) A schematic of a universal $8\times4$ optical linear multiplier with two unitary multipliers (red) consisting of MZIs in a grid-like layout and a diagonal layer (yellow). The MZIs of GridUnitary multipliers are indexed according to their layer depth (l) and dimension (d).  Symbols at the top represent the mathematical operations performed by the various modules. Inset: A MZI with two 50:50 beamsplitters and two tunable phaseshifters b) An FFT-like, non-universal multiplier with FFTUnitary multipliers (blue).}
    \label{fig:linear_layers}
\end{figure}

The ONN consists of multiple layers of programmable optical linear multipliers with intervening optical nonlinearities (Fig. \ref{fig:network_arch}). The linear multipliers are implemented with two unitary multipliers and a diagonal layer in the manner of a singular-value decomposition (SVD). These are, in turn, comprised of arrays of configurable MZIs, which each consist of two phaseshifters and two beamsplitters (Fig. \ref{fig:grid_arch}).
 
Complex-valued $N-$dimensional input vectors are encoded as coherent signals on $N$ waveguides. Unitary mixing between the channels is effected by MZIs and forms the basis of computation for ONNs. A single MZI consists of two beamsplitters and two phaseshifters (PS) (Fig. \ref{fig:grid_arch} inset). While the fixed 50:50 beamsplitters are not configurable, the two phaseshifters, parameterized by $\theta$ and $\phi$, are to be learned during training. Each MZI is characterized by the following transfer matrix (see App. \ref{app:transfer_matrix} for details):

\begin{align}
   U_{MZ}(\theta, \phi) =U_{BS} U_{PS}(\theta) U_{BS} U_{PS}(\phi) =
    i e^{i\theta/2}
   \begin{pmatrix}
    e^{i\phi}\sin\frac\theta2 & \cos\frac\theta2\\
    e^{i\phi}\cos\frac\theta2 & -\sin\frac\theta2
    \end{pmatrix}.
    \label{eq:U_MZI}
\end{align}

Early work has shown that universal optical unitary multipliers can be built with a triangular mesh of MZIs\cite{reck1994experimental}. These multipliers enabled the implementation of arbitrary unitary operations and were incorporated into the ONN design by Shen et al. \cite{shen2017deep}. Its asymmetry prompted the development of a symmetric grid-like network with more balanced loss\cite{clements2016optimal}. By relaxing the requirement on universality, a more compact design, inspired by the Cooley-Tukey FFT algorithm \cite{cooley1965algorithm}, has been proposed\cite{barak2007quantum}. It can be shown that FFT transforms, and therefore convolutions, can be achieved with specific phase configurations (see appendix \ref{app:FFT}). We allow the phase configurations to be learned for implementation of a greater class of transformations. 

In this section, we focus on the last two designs, referring to them as GridUnitary (Fig. \ref{fig:grid_arch}) and FFTUnitary (Fig. \ref{fig:fft_arch}), respectively. GridUnitary can implement unitary matrices directly by setting the phaseshifters using an algorithm by Clements et al.~\cite{clements2016optimal}. Despite being non-universal and lacking a decomposition algorithm, FFTUnitary can be used to reduce the depth of the unitary multipliers from $N$ to $\log_2(N)$. Reducing the number of MZIs leads to lower overall noise and loss in the network. However, due to the FFT-like design, waveguide crossings are necessary. To overcome this challenge, low-loss crossings\cite{ma2013ultralow} or 3D layered waveguides\cite{gattass2008femtosecond, panusa2019photoinitiator} could be utilized.

MZIs can also be used to attenuate each channel separately without mixing. This way, a diagonal multiplier can be built. Because signals can only be attenuated by MZIs, subsequent global optical amplification\cite{connelly2007semiconductor} is needed to emulate arbitrary diagonal matrices. Through SVD,
%singular value decomposition, 
a universal linear multiplier can be created from two unitary multipliers and a diagonal multiplier (Fig. \ref{fig:grid_arch}). Formally, a linear transformation represented by matrix $M$ can be decomposed as 
\begin{equation}
    M = \beta \cdot U \Sigma V^\dagger.
    \label{eq:SVD}
\end{equation}
Here both $U$ and $V^\dagger$ are unitary transfer matrices of GridUnitary multipliers while $\Sigma$ represents a diagonal layer with eigenvalues no greater than one. $\beta$ is a compensating scaling factor.

Along with linear multipliers, nonlinear layers are required for artificial neural networks. In fact, the presence of nonlinearties sets the study of ONNs apart from earlier research in linear photonic networks \cite{miller2017silicon}. One possible implementation is by saturable absorbers such as monolayer graphene \cite{bao2011monolayer}. This is has the advantage of being easily approximated with a Softplus function (see Sec. \ref{sec:software} for details on implementation). However, it has been demonstrated that Softplus underperforms, in many regards, when compared to rectified linear units (ReLU)\cite{nair2010rectified}. Indeed, a complex extension of ReLU, ModReLU, has been proposed \cite{arjovsky2016unitary}. While it is physically unrealistic to implement ModReLU, the nonoptimality of Softplus functions still motivates the exploration of other optical nonlinearities, such as optical bistability in microring resonators\cite{xu2006optical}, and two-photon absorption \cite{jiang2016analog, babaeian2018nonlinear} as alternatives.
 
\section{Neural network architecture and software implementation}
\label{sec:software}

\begin{figure}[h!]
    \centering
    \includegraphics[width=0.9\textwidth]{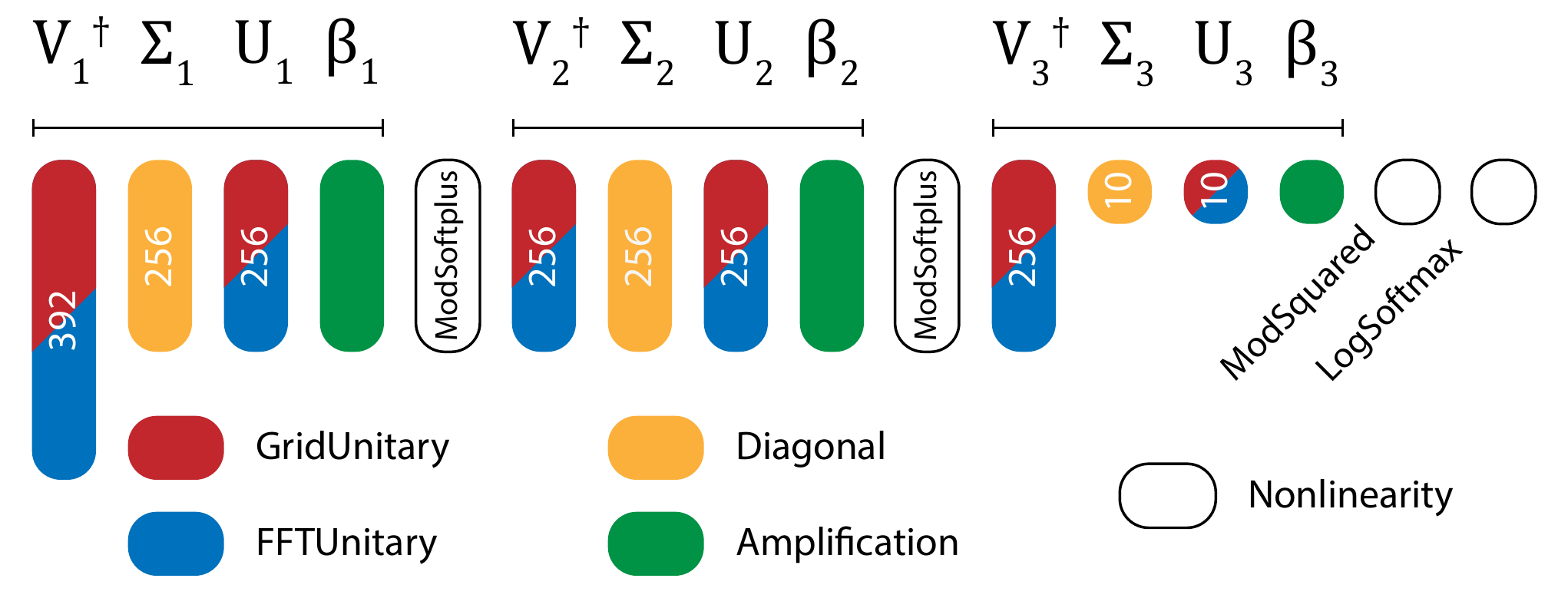}
    \caption{Network design used for the MNIST classification task. GridNet used universal unitary multipliers while FFTNet used FFT-Unitary multipliers. See Fig. \ref{fig:linear_layers} for details of physical implementation of the three linear layers.}
    \label{fig:network_arch}
\end{figure}
We considered a standard deep learning task of MNIST handwritten digit classification \cite{lecun1998mnist}. Fully connected feedforward networks with two hidden layers of 256 complex-valued neurons each were implemented with GridNet and FFTNet architectures (Fig. \ref{fig:network_arch}) and simulated in PyTorch \cite{paszke2017automatic}. The $28^2 = 784$ dimensional real-valued input was converted into $392 = 784/2$ dimensional complex-valued vectors by taking the top and bottom half of the image as the real and imaginary part. This was done to ensure the data is distributed evenly throughout the complex plane rather than just along the real number line.

Each network consists of linear multipliers followed by nonlinearities. The linear layers of GridNet and FFTNet were described in the previous section and illustrated in Fig. \ref{fig:linear_layers}. The response curve of the saturable absorption is approximated by the Softplus function\cite{dugas2001incorporating} (App. \ref{app:sat_abs}), a commonly used nonlinearity available in most deep learning libraries such as PyTorch. The nonlinearity is applied to the modulus of the complex numbers. A modulus squared nonlinearity modeling an intensity measurement is then applied. The final SoftMax layer allows the (now real) output to be interpreted as a probability distribution. A cross-entropy\cite{Cover:2006:EIT:1146355} loss function is used to evaluate the output distribution against the ground truth.
 
An efficient implementation of GridNet requires representing matrix-vector multiplications as element-wise vector multiplications \cite{jing2017tunable}. Nevertheless, training the phaseshifters directly was still time consuming. Instead, a complex-valued neural network \cite{trabelsi2017deep} was first trained. An SVD (Eq. \eqref{eq:SVD}) was then performed on each complex matrix. Finally, phaseshifters were set to produce the unitary ($U, V^\dagger$) and diagonal ($\Sigma$) multipliers through a decomposition scheme by Clements et al. \cite{clements2016optimal}.

However, note that SVD is ambiguous up to permutations ($\Pi$) of the singular values and the columns of $U$ and $V$.
\begin{align}
    U\Sigma V^\dagger = (U \Pi^{-1}) (\Pi \Sigma \Pi^{-1}) (\Pi V^\dagger).
    \label{eq:svd_perm}
\end{align}
Conventionally, the ambiguity is resolved through ordering the singular values from largest to smallest. In Sec. \ref{sec:pos_sen} we show that randomizing the singular values increases the error tolerance of GridNet. FFTNet is trained directly and its singular values are naturally unordered. For a fair comparison, we randomly permute the singular values of GridNet.

After 10 training epochs with standard stochastic gradient descent\cite{robbins1985stochastic}, classification accuracies of $97.8\%$ (GridNet) and $94.8\%$ (FFTNet) were achieved. Better accuracies can be achieved through convolutional layers \cite{simard2003best}, Dropout regularization\cite{srivastava2014dropout}, better training methods, etc. However, we omitted these in order to focus purely on the effects of architecture.

The networks were trained assuming ideal components represented with double-precision floating point values. Under realistic conditions, due to imprecision in fabrication, calibration, etc., the realizable accuracy could be much lower. During inference, we modeled these imprecisions by adding independent zero-mean Gaussian noise of standard deviation $\sigma_{PS}$ and $\sigma_{BS}$ to the phases $(\theta, \phi)$ of the phaseshifters and the transmittance $T$ of the beamsplitters, respectively. Reasonable values for such imprecisions can be taken to be approximately $\sigma_{PS} \approx 0.01\text{rad}$ and $\sigma_{BS} \approx 1\% = 0.01$ \cite{flamini2017benchmarking, flamini2015thermally}. Note that the dynamical variation due to laser phase noise can be modeled by $\sigma_{PS}$ as well. However, we show in App. \ref{app:laser_phase_noise} that typical values would be well below $0.01$ rad. 

\section{Results}
\subsection{Degradation of network accuracy}
\label{sec:accuracy}

\begin{figure}[ht!]
    \centering
    \subfloat[Ideal GridNet]{%
    \label{fig:ideal_grid}%
    \includegraphics[width=0.25\columnwidth]{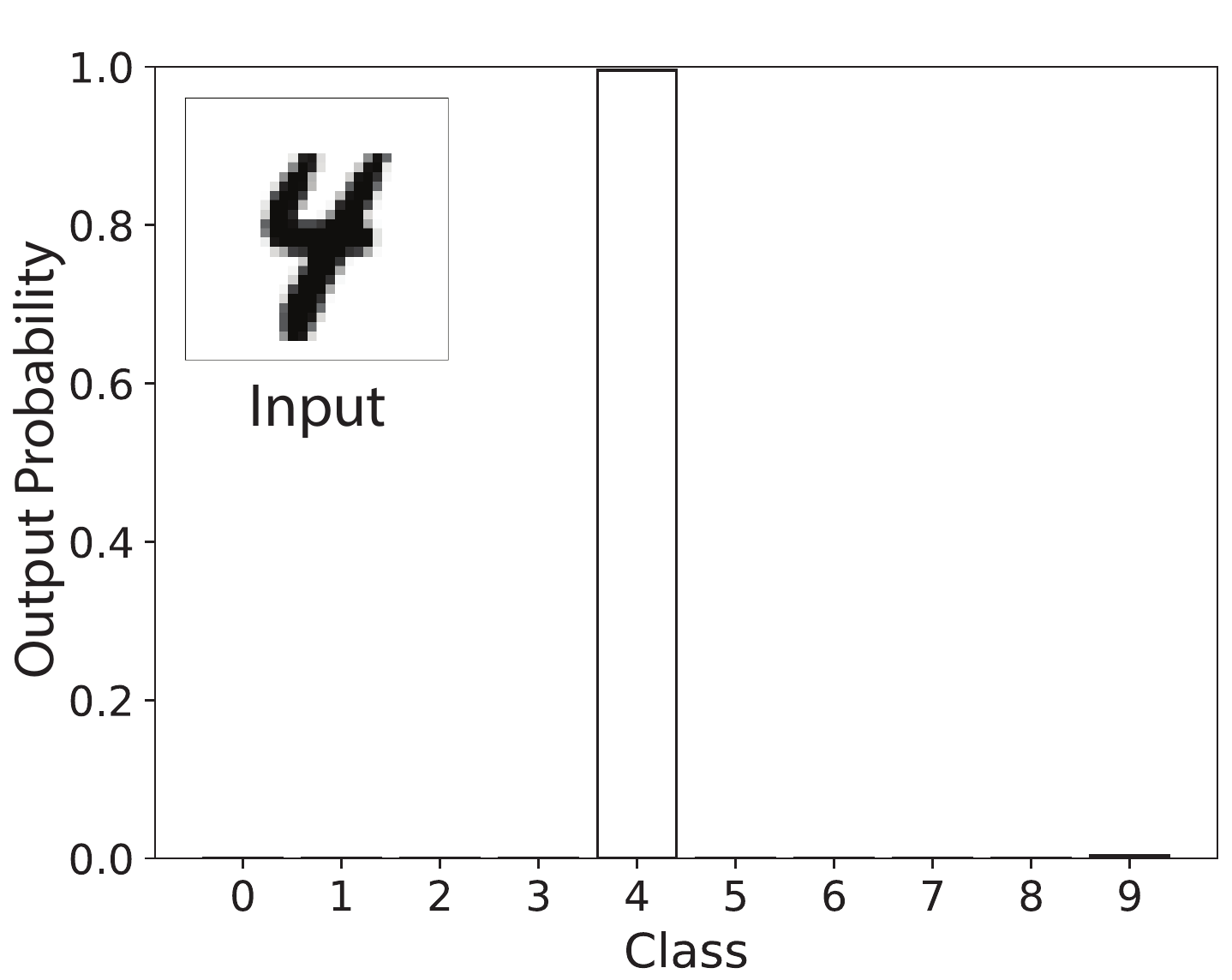}}%
    \subfloat[Imprecise GridNet]{%
    \label{fig:imprecise_grid}%
    \includegraphics[width=0.25\columnwidth]{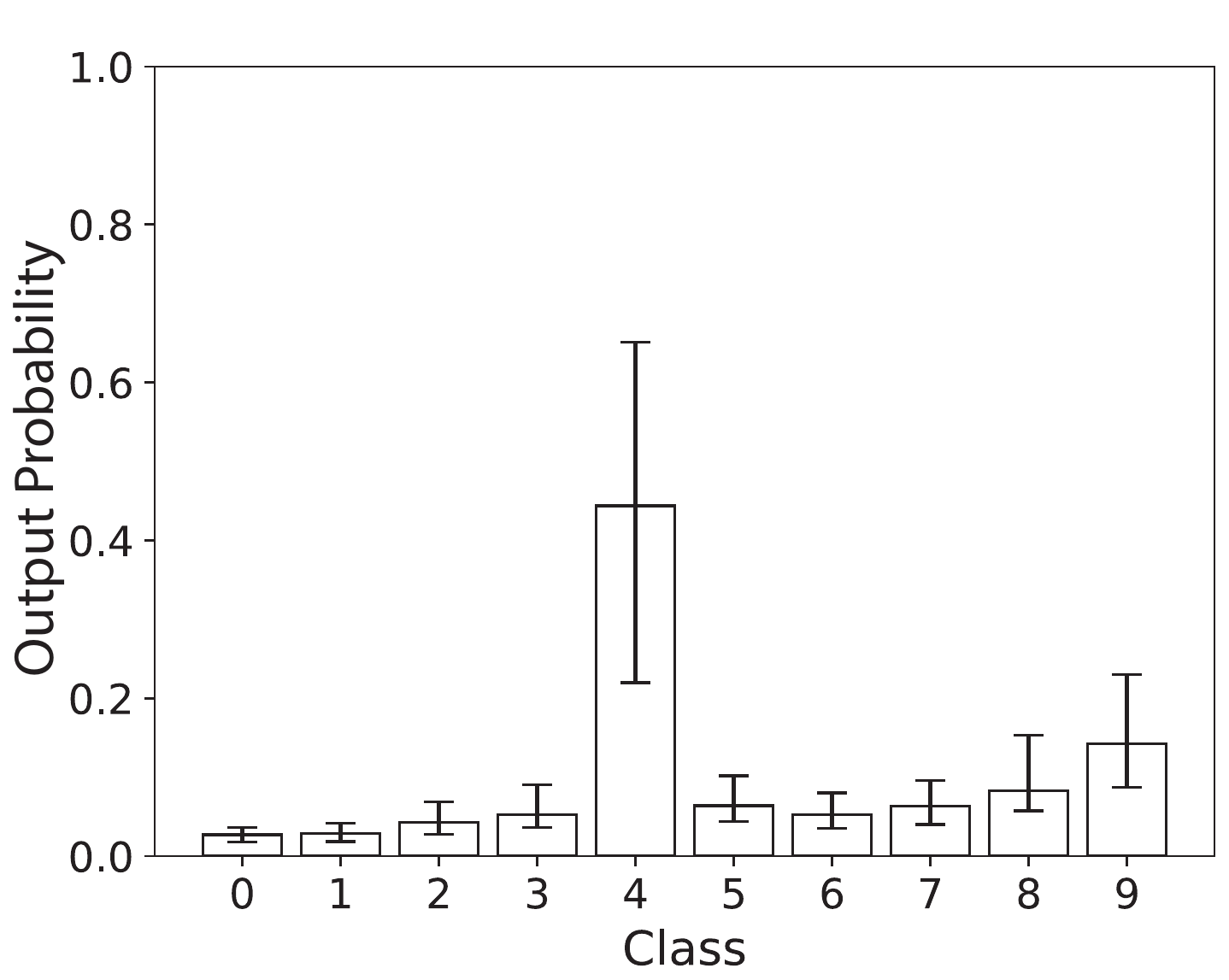}}%
    \subfloat[Ideal FFTNet]{%
    \label{fig:ideal_FFT}%
    \includegraphics[width=0.25\columnwidth]{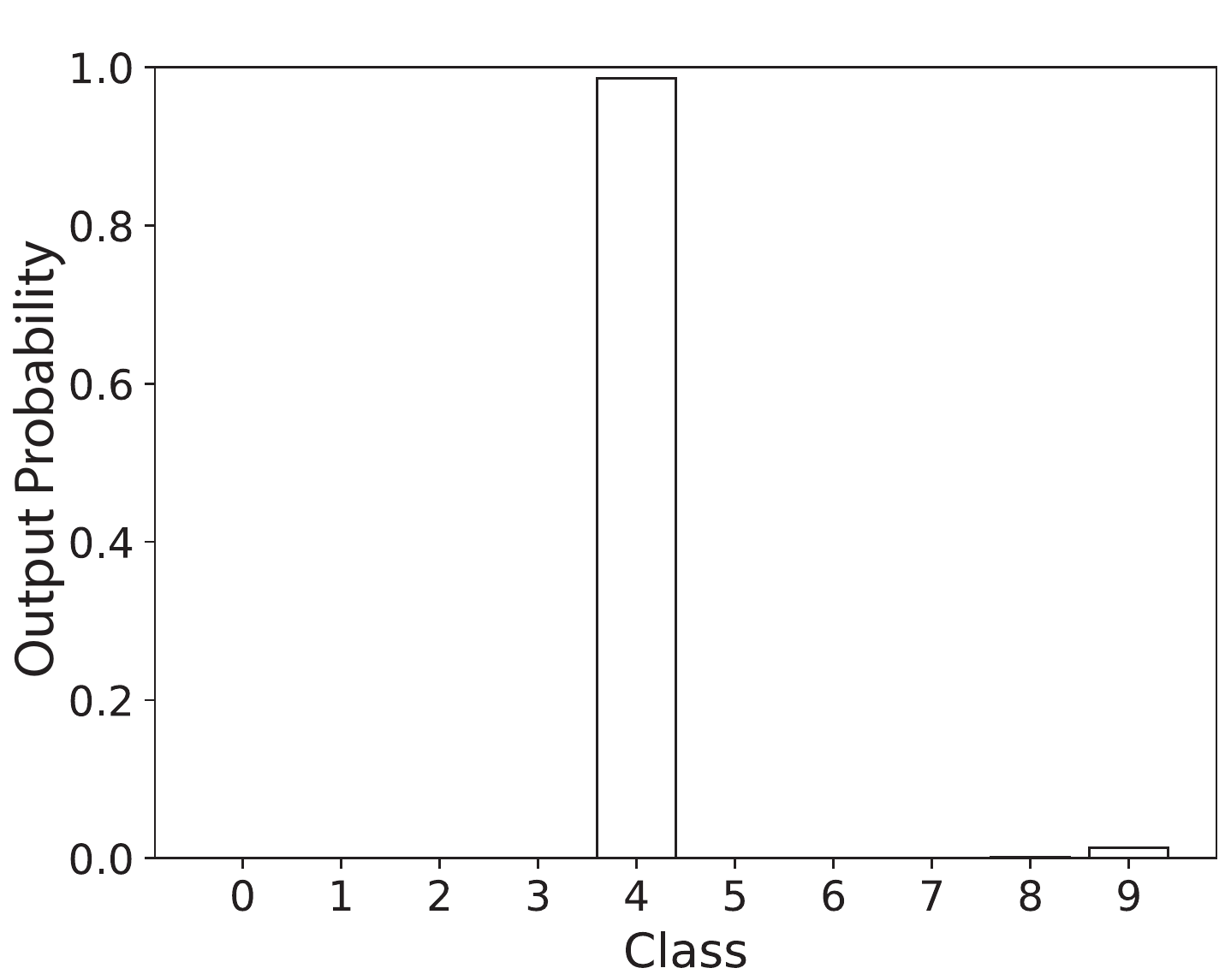}}%
    \subfloat[Imprecise FFTNet]{%
    \label{fig:imprecise_FFT}%
    \includegraphics[width=0.25\columnwidth]{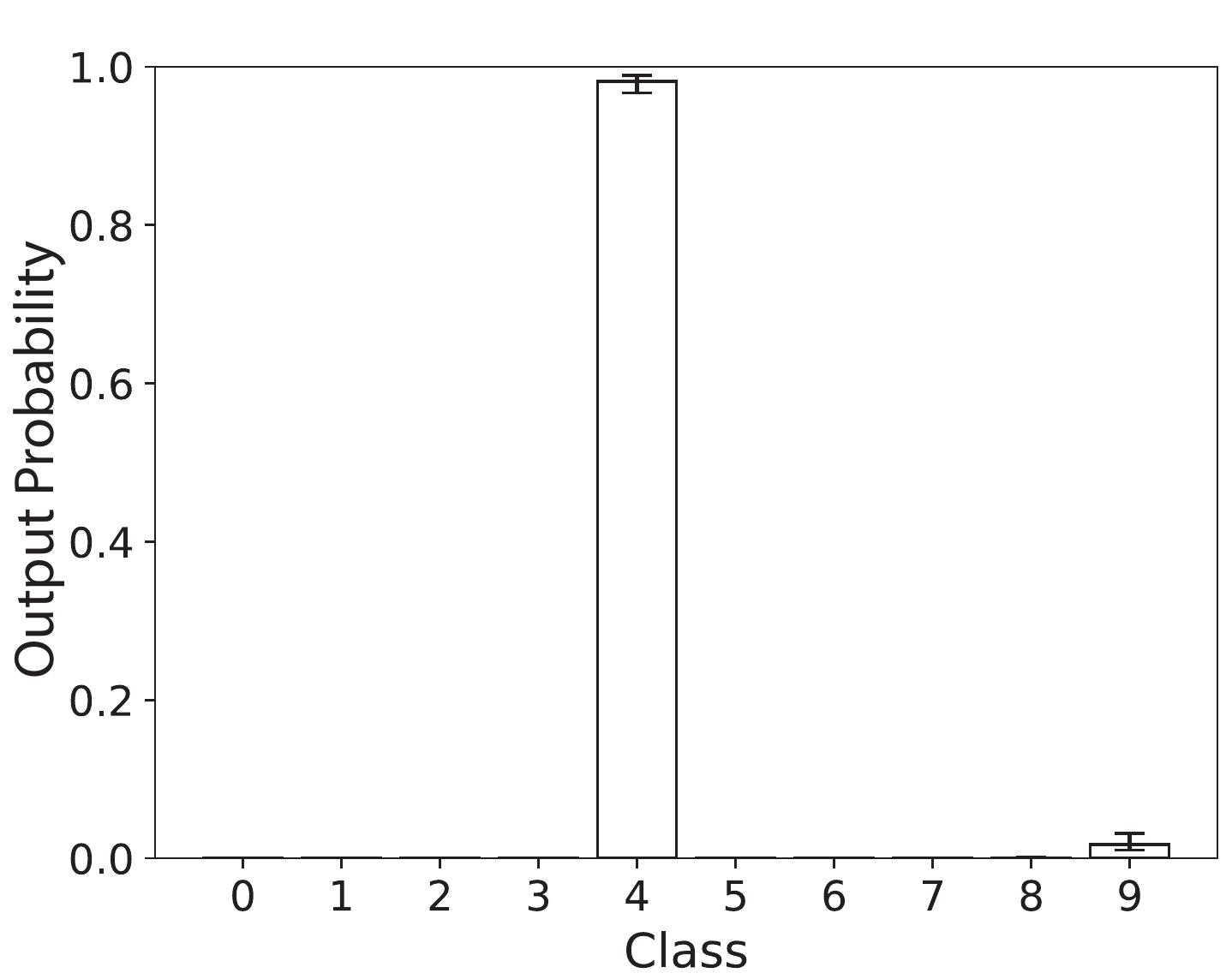}}%
    \caption{Visualizing the degradation of ONN outputs, FFTNet is seen to be much more robust than GridNet. Identical input is fed through GridNet (a, b) and FFTNet (c, d), simulated with ideal components (a, c) and imprecise components (b, d) with $\sigma_{BS} = 0.01$ and $\sigma_{PS} = 0.01\text{rad}$. Imprecise networks are simulated 100 times and their mean output is represented by bar plots. Error bars represent the 20th to 80th percentile range.}
    \label{fig:output_degrade}
\end{figure}

To investigate the degradation of the networks due to imprecisions, we started by simulating 100 instances of imprecise networks with $\sigma_{BS} = 1\%$ and $\sigma_{PS} = 0.01\text{rad}$. Identical inputs of a digit ``4'' (Fig. \ref{fig:ideal_grid} inset) are fed through each network. The mean and spread of the output of the ensemble is plotted and compared against the output from the ideal network (Fig. \ref{fig:output_degrade}). 

The degradation of classification output is significant for GridNet. Without imprecisions in the photonic components, the digit is correctly classified with near 100\% confidence (Fig. \ref{fig:ideal_grid}). When imprecisions are simulated, we see a large decrease in classification confidence (Fig. \ref{fig:imprecise_grid}). In particular, the image is often misclassified when the prediction probability for class ``9'' is greater than that for class ``4''. Repeating these experiments on FFTNet demonstrated that they were much more resistant to imprecisions (Fig. \ref{fig:ideal_FFT}, \ref{fig:imprecise_FFT}). In Appendix \ref{app:conf_mat}, we show confusion matrices of both networks with increasing error to further support this conclusion.
\begin{figure}[h!]
    \centering
    \subfloat[GridNet]{%
    \label{fig:grid_acc}%
    \includegraphics[width = 0.5 \columnwidth]{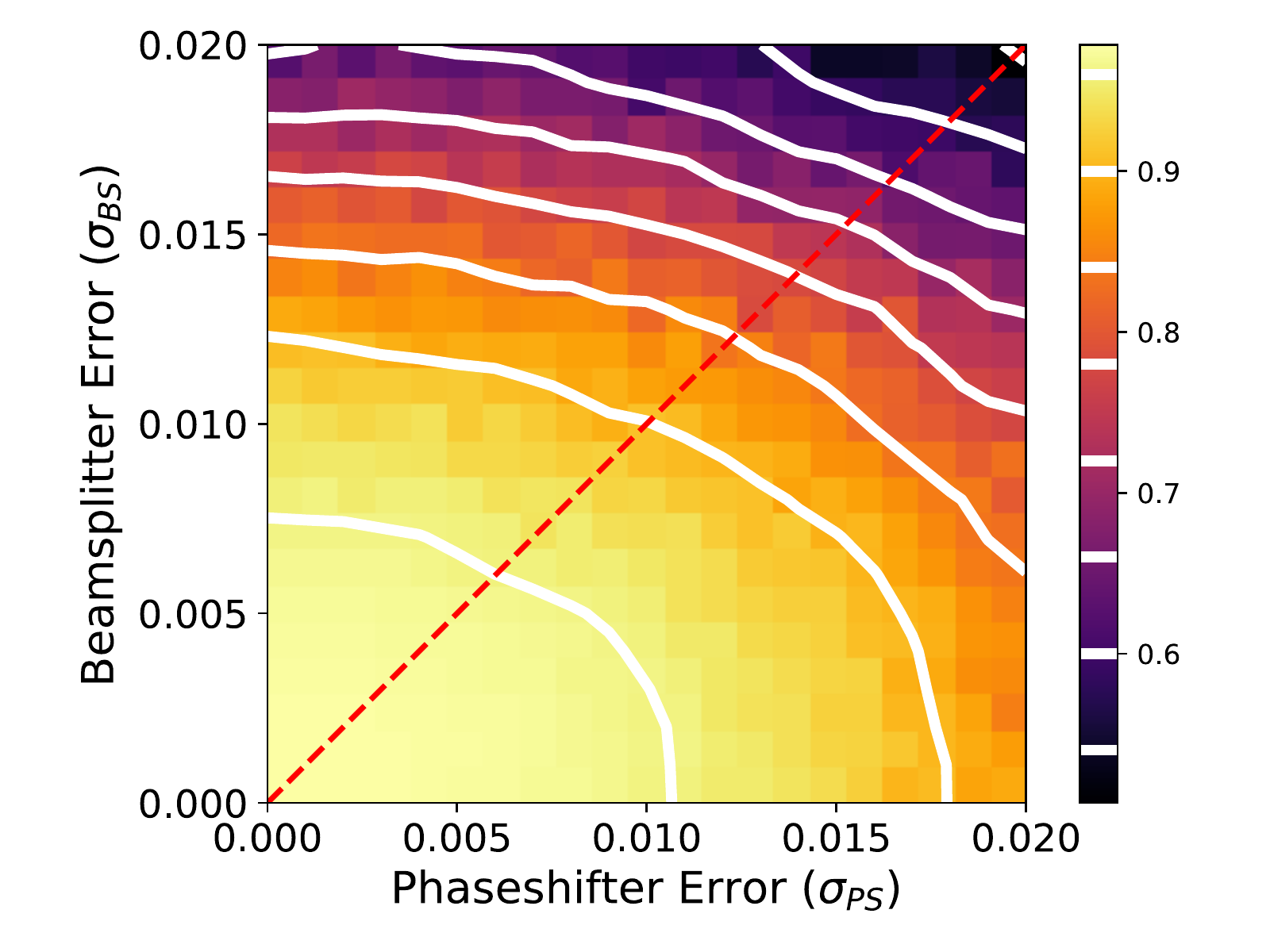}}%
    \subfloat[FFTNet]{%
    \label{fig:fft_acc}%
    \includegraphics[width = 0.5 \columnwidth]{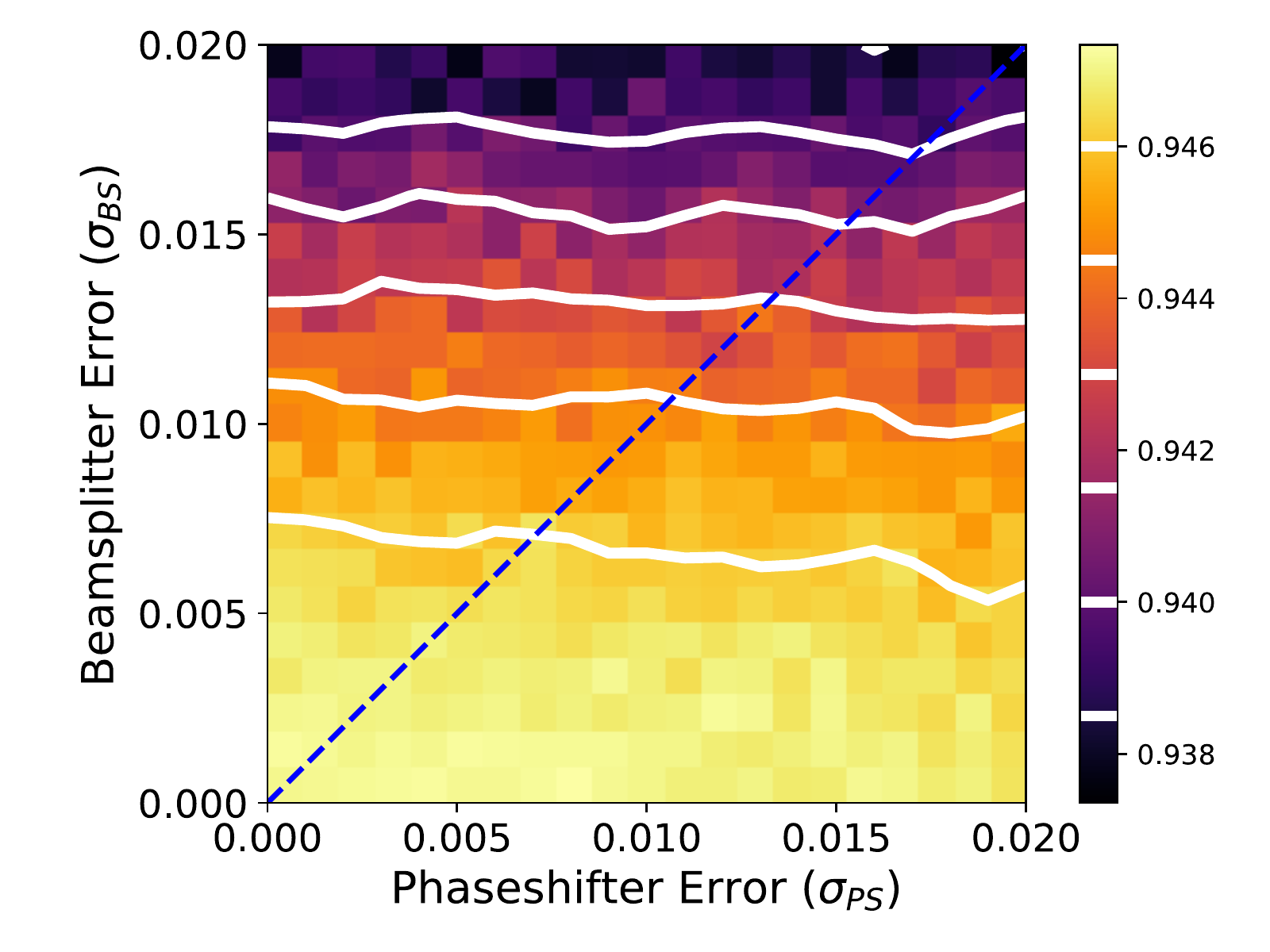}}%
    \\
    \subfloat[Direct Comparison of GridNet and FFT Accuracy]{%
    \label{fig:grid_vs_fft}%
    \includegraphics[width = 0.5 \columnwidth]{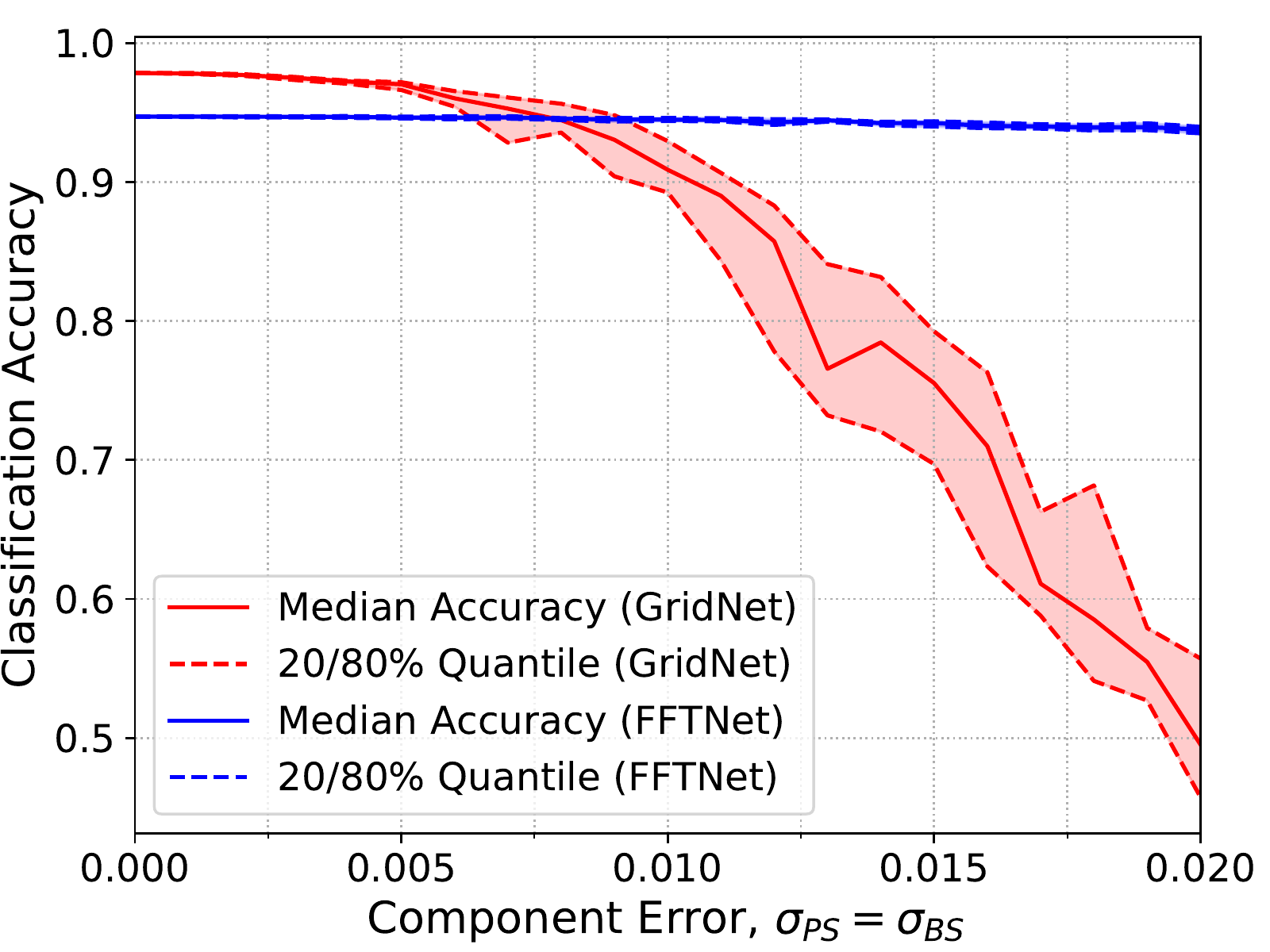}}%
    \caption{The decrease in classification accuracy is visualized for GridNet and FFTNet. (a,b) The two networks were tested with simulated noise of various levels for 20 runs. The mean accuracy is plotted as a function of $\sigma_{PS}$ and $\sigma_{BS}$. Note the difference in color map ranges between the two plots. (c) The accuracies of GridNet and FFTNet are compared along the $\sigma_{PS} = \sigma_{BS}$ cutline.}
    \label{fig:acc_compare}
\end{figure}

Evaluating the two networks on overall classification accuracy confirms the superior robustness to imprecisions of FFTNet. GridNet and FFTNet were tested at levels of imprecisions with of imprecisions with $\sigma_{PS}/\text{rad}$ and $\sigma_{BS}$ ranging from $0$ to $0.02$ with a step size of $0.001$. At each level of imprecision, 20, instances of each network were created and tested. The mean accuracies are plotted in Fig. \ref{fig:grid_acc}, \ref{fig:fft_acc}. A direct comparison between the two networks along the diagonal (i.e., $\sigma_{PS} = \sigma_{BS}$ cut line, taking $1\% = 0.01$ rad) is shown in Fig. \ref{fig:grid_vs_fft}.

Starting at roughly 98\% with ideal components, the accuracy of GridNet rapidly drops with increasing $\sigma_{PS}$ and $\sigma_{BS}$. By comparison, very little change in accuracy is seen for FFTNet despite starting with a lower ideal accuracy. Also of note are the qualitatively different levels of sensitivity of the different components to imprecision. In particular, FFTNet is much more resistant to phaseshifter error compared to beamsplitter error.

The experiments described in this section confirm the significant effect component imprecisions have on the overall performance of ONNs, as well as the importance of architecture in determining the network's robustness of the network to these imprecisions. Despite having a better classification accuracy in the absence of imprecisions, GridNet is surpassed by FFTNet when a small amount of error ($\sigma_{PS} = 0.01, \sigma_{BS}$ = 1\%rad) is present. In Appendix \ref{app:quant_err}, we demonstrate that FFTNet is also more robust to quantization error that GridNet.

%==========Stacked and Truncated==============
\subsection{Stacked FFTUnitary and truncated GridUnitary}
\label{sec:stack_trunc}
\begin{figure}[h!]
    \centering
    \subfloat[StackedFFT and FFTUnitary]{%
    \label{fig:stacked_fft}%
    \includegraphics[width = 0.4 \columnwidth]{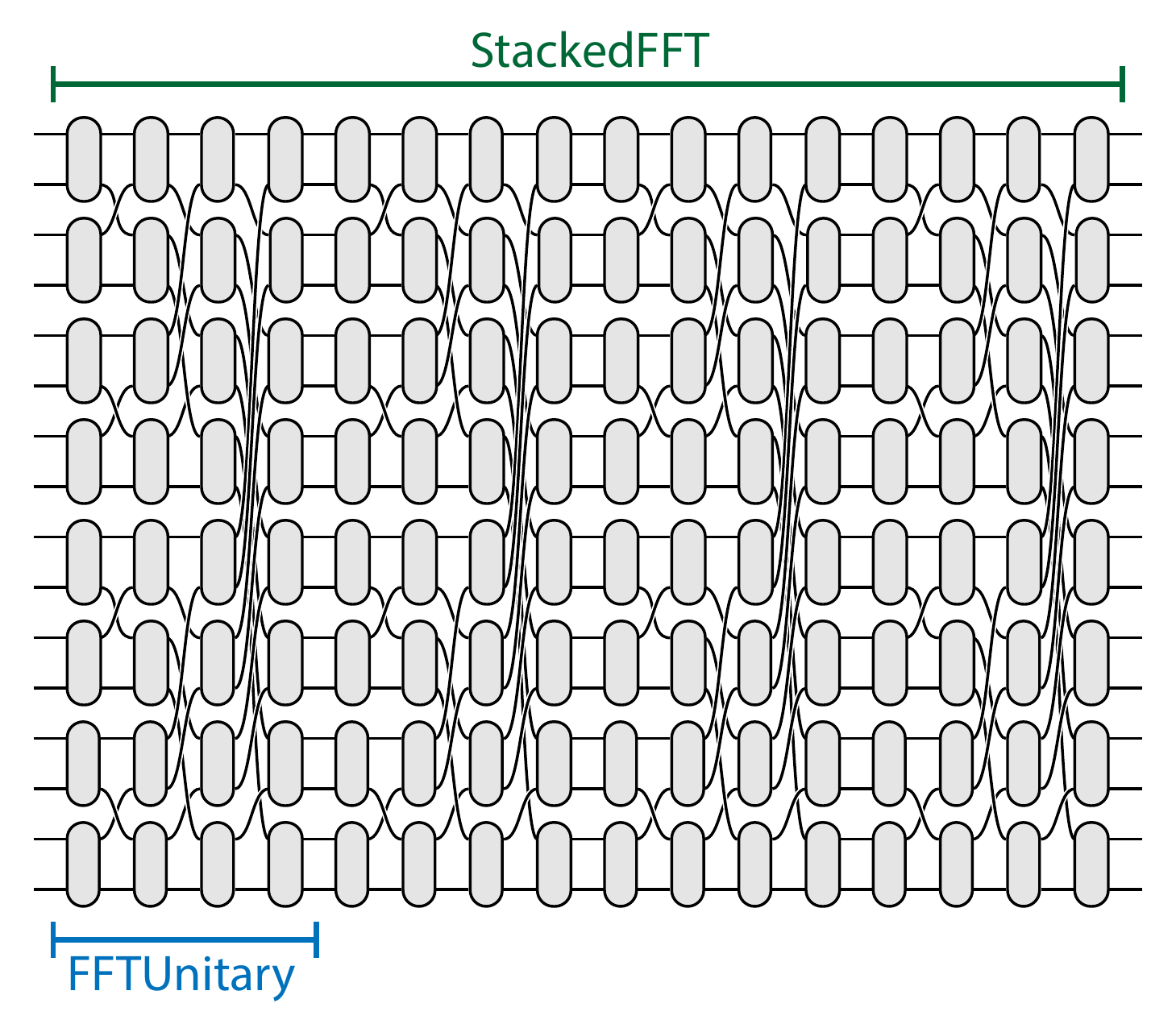}}%
    \subfloat[TruncGrid and GridUnitary]{%
    \label{fig:trunc_grid}%
    \includegraphics[width = 0.4 \columnwidth]{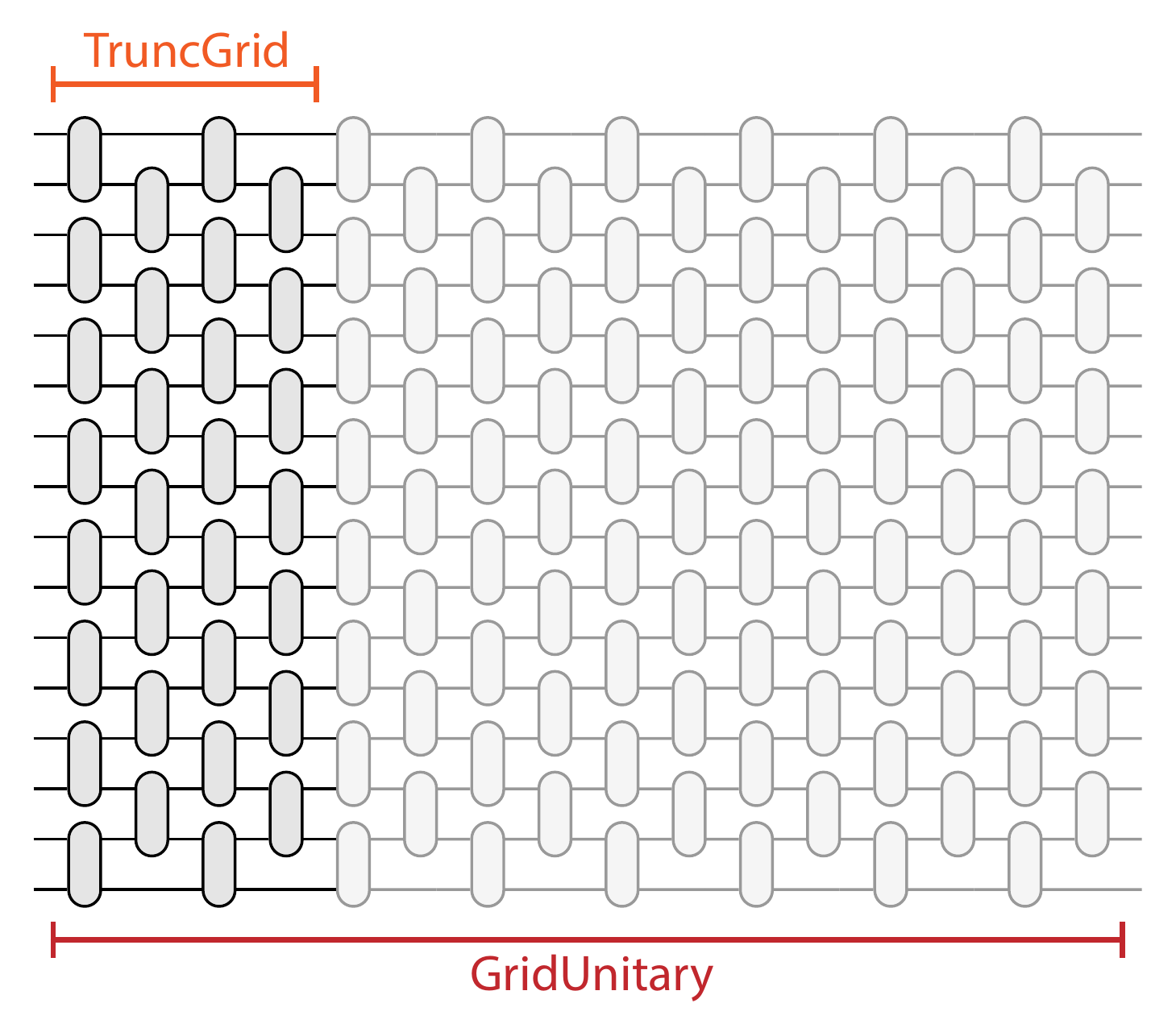}}%
    \label{fig:stack_trunc}
    \caption{The architecture of a) StackedFFT and b) TruncGrid shown with FFTUnitary and GridUnitary from which they were derived. For clarity, the dimension, here, is $N=2^4=16$ so FFTUnitary was stacked four times and GridUnitary was truncated at the fourth layer. In the experiments described in this section, the dimension was taken to be $N=2^8=256$.}
\end{figure}

One obvious reason why FFTNet would be more robust than GridNet is its much lower number of MZI layers. Their respective, constituent unitary multipliers, FFTUnitary and GridUnitary contains $\log_2(N)$ and $N$ layers respectively. For $N = 2^8 = 256$, GridUnitary is $32$ times deeper than FFTUnitary which contains only $8$ layers.

To demonstrate that FFTUnitary is more robust due architectural reasons beyond its shallow depth, in this section, we introduce two unitary multipliers -- StackedFFT (Fig. \ref{fig:stacked_fft}) and TruncGrid (Fig. \ref{fig:trunc_grid}). StackedFFT consists of FFTUnitary multipliers stacked end-to-end 32 times and TruncGrid is the GridUnitary truncated after 8 layers of MZIs. This way, FFTUnitary and TruncGrid have the same depth as do GridUnitary and StackedFFT.

\begin{figure}[h!]
    \centering
    \subfloat[StackedFFT and GridUnitary]{%
    \label{fig:stacked_grid}%
    \includegraphics[width = 0.4 \columnwidth]{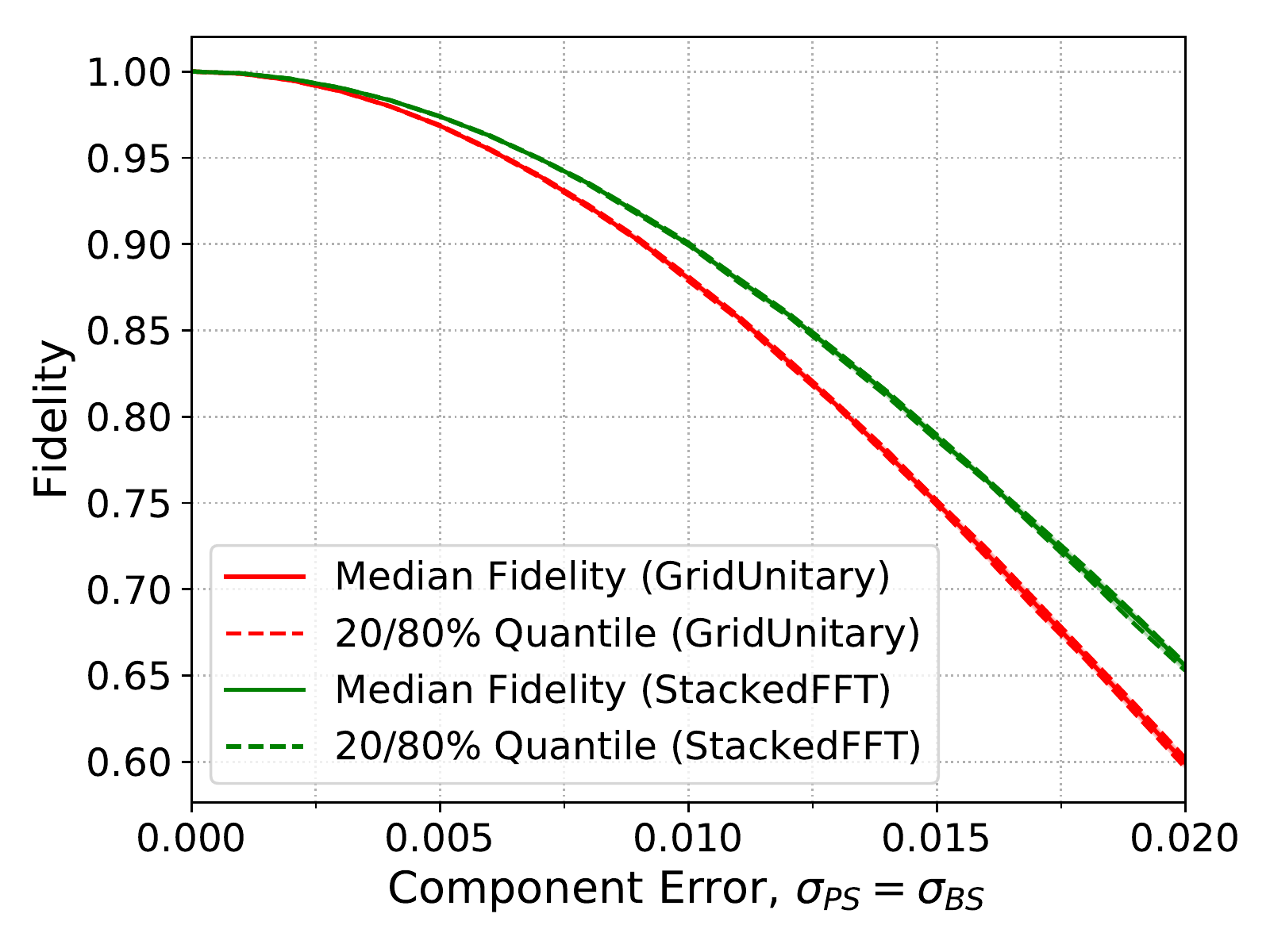}}%
    \subfloat[TruncGrid and FFTUnitary]{%
    \label{fig:trunc_fft}%
    \includegraphics[width = 0.4 \columnwidth]{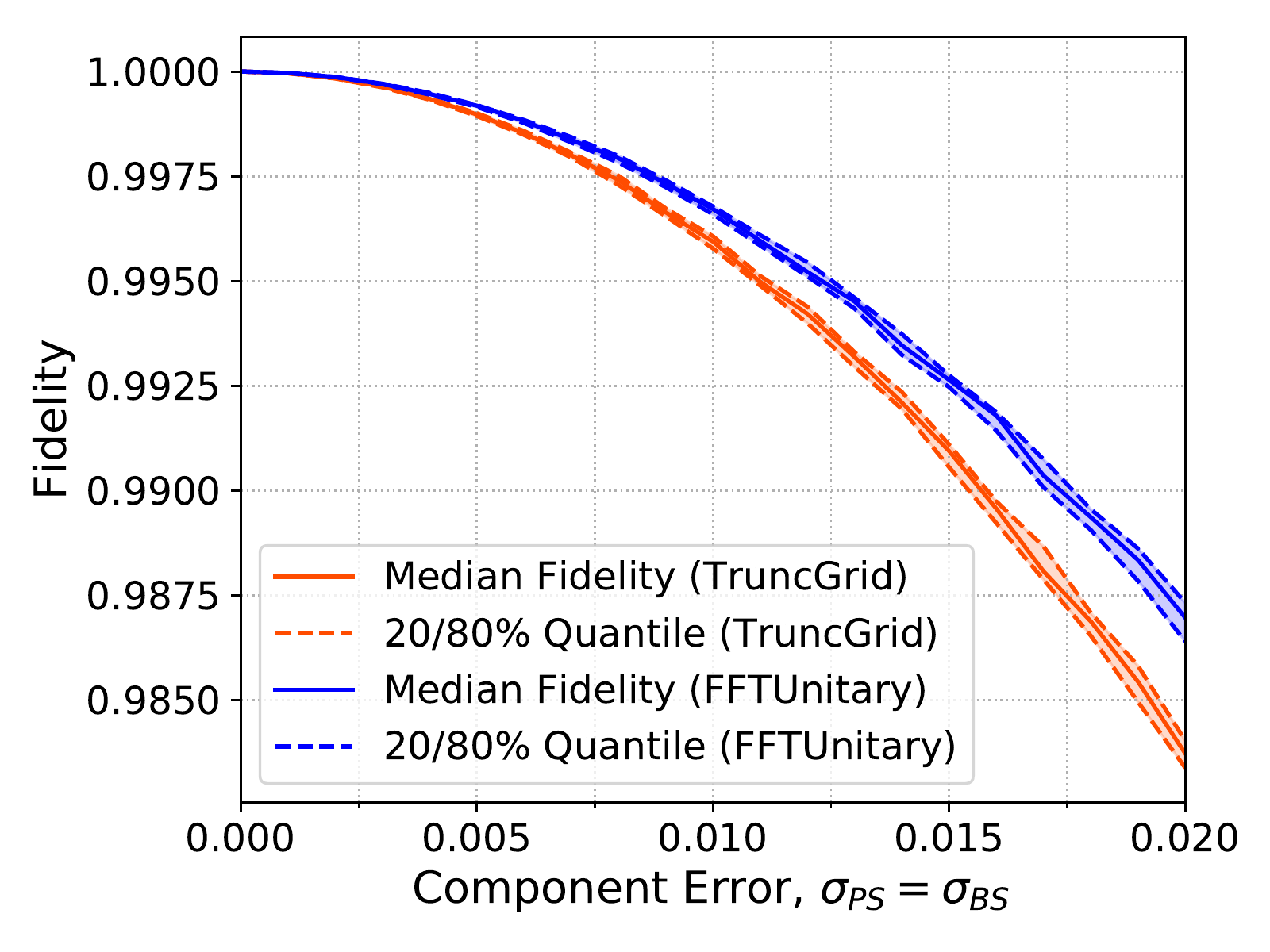}}%
    \label{fig:stack_trunc_fid}
    \caption{With the same layer depth, multipliers with FFT-like architectures are shown to be more robust. The fidelity between the error-free and imprecise transfer matrices is plotted as a function of increasing error. Two sets of comparisons between unitary multipliers of the same depth are made. a) Both StackedFFT and GridUnitary have $N = 256$ layers of MZIs. b) TruncGrid and FFTNet have $\log N = 8$ layers.}
\end{figure}
Unitary multipliers by themselves are not ONNs and cannot be trained for classification tasks. Instead, after introducing imprecisions to the each multiplier, we evaluated the fidelity $F(U_0, U)$ between the original, error-free transfer matrix $U_0$ and the imprecise transfer matrix $U$. The fidelity, a measure of ``closeness'' between two unitary matrices, is defined as\cite{walls2007quantum}
\begin{align}
    F(U_0, U) = \left|\frac{\Tr(U^\dagger U_0)}{N} \right|^2.
\end{align}
Ranging from 0 to 1, $F(U_0, U) = 1$ only when $U = U_0$. Using this metric of fidelity, we show that StackedFFT is more robust to error than GridUnitary (Fig. \ref{fig:stacked_grid}) and TruncGrid more than FFTUnitary (Fig. \ref{fig:trunc_fft}). Both comparisons are between multipliers with the same number of MZI layers. Yet, the FFT-like architectures are still more robust to their grid-like counterparts.

One possible explanation could be the better mixing facilitated by FFTUnitary. GridUnitary and thus TruncGrid, at each MZI layer, only mixes neighboring waveguides. After $P$ layers, each waveguide is connected to, at most, to its $2P$ nearest neighbors. In comparison, after $P$ layers, FFTUnitary connects $N = 2^P$.

Here, we have compared the robustness of different unitary multipliers in isolation. We stress that the overall robustness of neural networks is a much more complex and involved problem. A rough understanding can be formulated as follows. A trained neural network defines a decision boundary throughout the input space. Introduction of errors perturbs the decision boundary which can lead to misclassification. To reduce this effect, we can make the decision boundary of ONNs more robust to errors. However, it is also important to consider the robustness of misclassification due to perturbations of decision boundaries. Indeed, it has been shown that robustness of neural networks are dependent on the geometry of the boundary \cite{fawzi2016robustness}. 

A complete analysis of the robustness of neural networks to various forms of perturbations is outside the scope of this paper. Nonetheless, it is important to understand the dependence of ONNs on both architectural and algorithmic design. 

%(see Fig. \subref*{fig:transfer_matrices}).
\iffalse
\begin{figure}[h!]
    \centering
    \subfloat[TruncGrid Transfer Matrix]{%
    \label{fig:U_trunc}%
    \includegraphics[width = 0.4 \columnwidth]{U_trunc.pdf}}%
    \subfloat[FFTUnitary Transfer Matrix]{%
    \label{fig:U_fft}%
    \includegraphics[width = 0.4 \columnwidth]{U_fft.pdf}}%
    \caption{With the same number of MZI layers, FFTUnitary can be seen to mix signals more efficiently than TruncGrid. The absolute value of the transfer matrices of both unitary multipliers are plotted. Here, the dimension is $N=2^4=16$ and TruncGrid was truncated at layer $4$.}
    \label{fig:transfer_matrices}
\end{figure}
\fi

\subsection{Localized imprecisions}
\label{sec:pos_sen}
\begin{figure}[h!]
    \centering
    \includegraphics[width=\columnwidth]{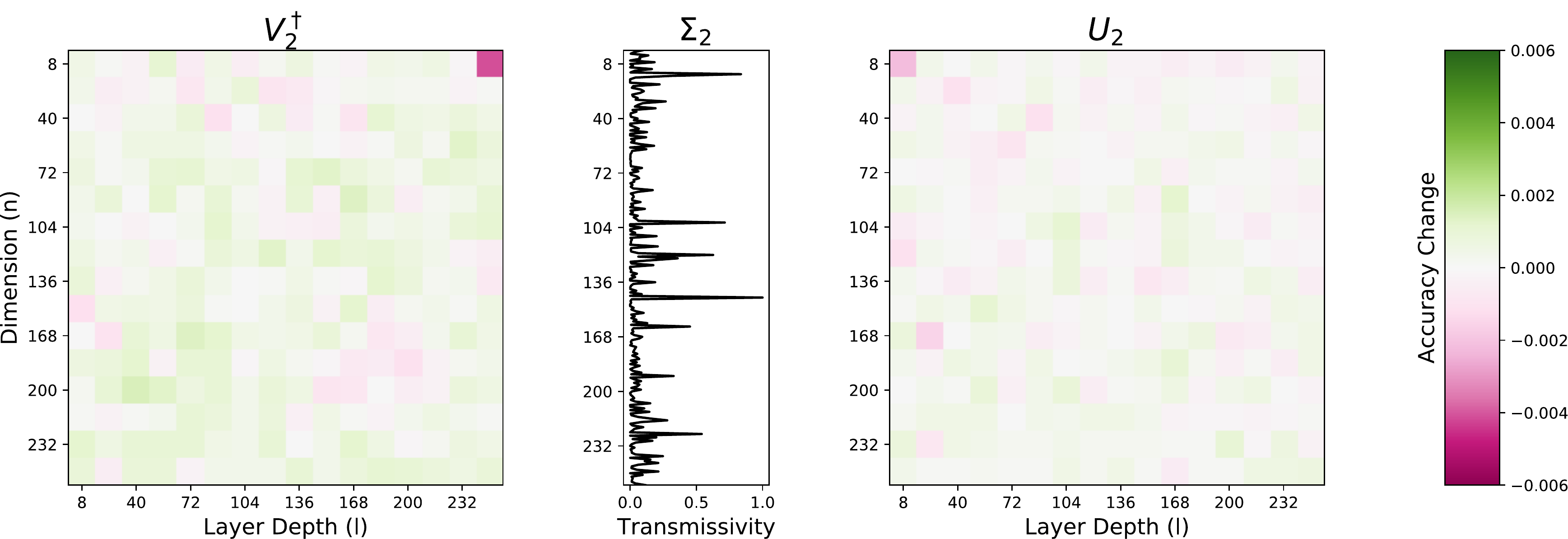}
    \caption{Change in accuracy due to localized imprecision in layer 2 of GridNet with randomized singular values. A large amount of imprecision ($\sigma_{PS} = 0.1\text{rad}$) is introduced to $8 \times 8$ blocks of MZIs in an otherwise error-free GridNet. The resulting change in accuracy of the network is plotted as a function of the position of the MZI block in GridUnitary multipliers $V_2^\dagger$ and $U_2$ (coordinates defined as in Fig. \ref{fig:grid_arch}). The transmissivity of each waveguide through the diagonal layer $\Sigma_2$ is also plotted (center panel).}
    \label{fig:pos_sen_random}
\end{figure}
To better understand the degradation of network accuracy, we mapped out the sensitivity of GridNet to specific groups of MZIs. A relatively large amount of imprecision ($\sigma_{PS} = 0.1\text{rad}$) was introduced to $8 \times 8$ blocks of MZIs in layer 2 (Fig. \ref{fig:network_arch}) of an otherwise error-free GridNet. The resulting change in classification accuracy is plotted as a function of the position of the MZI block (Fig. \ref{fig:pos_sen_random}). We see no strong correlation between the change in accuracy and the spatial location of the introduced error. In fact, error in many locations led to small increases in accuracy, 
%implying 
suggesting that much of the effect is due to chance.

This result seems to contradict previous studies on the spatial tolerance of MZIs in a GridUnitary multiplier \cite{pai2018matrix,russell2017direct,burgwal2017using}. It was discovered that the central MZIs of the multiplier had a much lower tolerance than those near the edges. When learning randomly sampled unitary matrices, the central MZIs needed to have phase shift values very close to $0$ ($\pi$, following the convention used in this paper). This would only be achievable with MZIs with extremely high extinction ratios and thus low fabrication error.

Empirically, this distribution of phases was observed in GridUnitary multipliers of trained ONNs (See app. \ref{app:phase_distr}). However, the idea of tolerance of a MZI to beamsplitter fabrication imprecision, while related, is not the same as the network sensitivity to localized imprecisions. To elaborate, tolerance is implicitly defined, in references\cite{pai2018matrix,russell2017direct,burgwal2017using}, as roughly the allowable beamsplitter imperfection (deviation from 50:50) that still permits post-fabrication optimization of phaseshifter towards arbitrary unitary matrices. In our pre-fabrication optimization approach, we take sensitivity to be the deviation from ideal classification accuracy when imprecision is introduced to the MZI with no further reconfiguration. See App. \ref{app:hybrid} for this difference further illustrated by experiments with another architecture.

\begin{figure}[h!]
    \centering
    \includegraphics[width=\columnwidth]{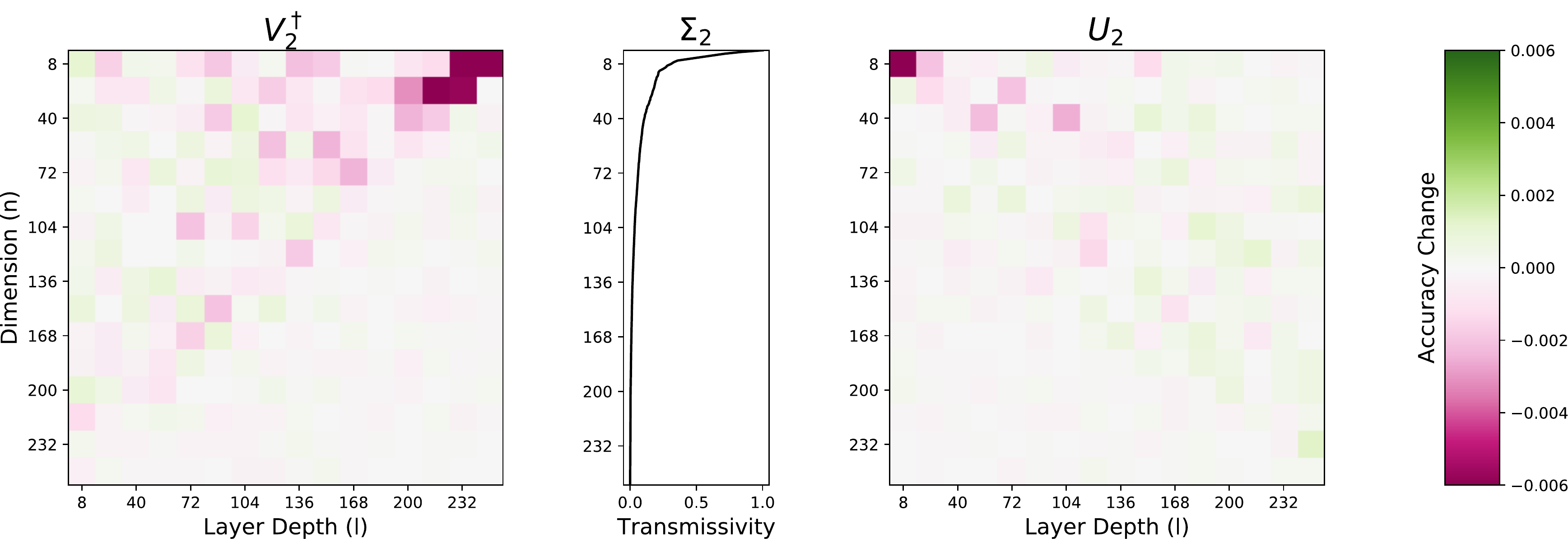}
    \caption{Effects of localized imprecision in layer 2 of GridNet with ordered singular values. Similar to Fig. \ref{fig:pos_sen_random}, except GridNet has its singular values ordered. Therefore, the transmissivity is also ordered (center panel).}
    \label{fig:pos_sen_order}
\end{figure}
Recall that the singular values $\Sigma$ of the GridNet's linear layers could be permuted together with columns and rows of $U$ and $V^\dagger$ respectively without changing the final transfer matrix (Eq. \eqref{eq:svd_perm}). The singular values were randomized to provide a fair comparison with FFTNet. We then performed the same experiment on GridNet where the singular values of each layer were not randomized but ordered from largest to smallest. Therefore, the transmissivity $T = |\sin(\theta/2)|^2$ of the diagonal multiplier $\Sigma$ is also ordered (Fig. \ref{fig:pos_sen_order}). In this case, there is a significant, visible pattern because most of the signal travels through the top few waveguides of $\Sigma_2$ due to the ordering of transmissivities. Only MZIs connected to those waveguides have a strong effect on the network. In fact, the network is especially sensitive to imprecisions in MZIs closest to this bottleneck (Fig. \ref{fig:pos_sen_order}, top-right of $V_2^\dagger$ and top-left of $U_2$). It is important to note that this bottleneck only exist due to the locality of connections in GridNet where only neighboring waveguides are connected by MZIs. In FFTNet, due to crossing waveguides, no such locality exist.

\begin{figure}[h!]
    \centering
    \includegraphics[width=0.5\columnwidth]{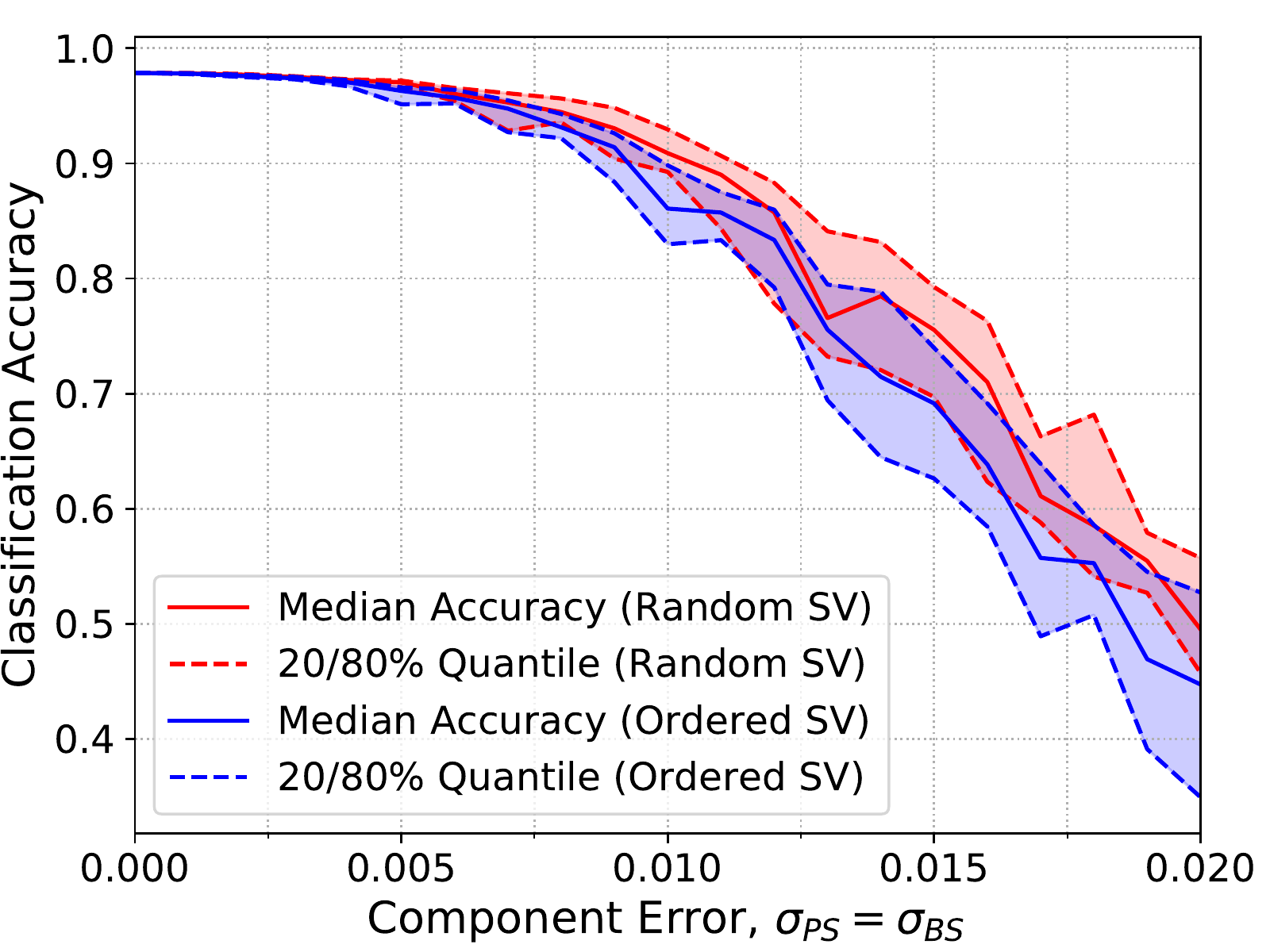}
    \caption{The degradation of accuracies with increasing $\sigma_{PS} = \sigma_{BS}$ compared between two GridNets one with ordered and another with randomized (but fixed) singular values.}
    \label{fig:compare_order}
\end{figure}

In addition to, and likely due to the spatial non-uniformity in error sensitivity, GridNet with ordered singular values is more susceptible to uniform imprecisions (Fig. \ref{fig:compare_order}). The same GridNet architecture, could be made more resistant by shuffling its singular values. This difference between two identical architectures implementing identical linear and non-linear transformations demonstrates that the resistance to error in ONNs is effected by more than architecture.

\section{Conclusion}
Having argued that pre-fabrication, software optimization of ONNs is much more scalable than post-fabrication, on-chip optimization, we compared two types of networks--GridNet and FFTNet in their robustness to error. These two networks were selected to showcase the trade-off between expressivity and robustness. We demonstrated in Sec. \ref{sec:accuracy} that the output of GridNet is much more sensitive to errors than FFTNet. We have illustrated the robustness of FFTNet by a providing a thorough evaluation of both networks operating with imprecisions ranging between $0 \le \sigma_{BS}, \sigma_{PS} \le 0.02$. With ideal accuracies of $97.8\%$ and $94.8\%$ for GridNet and FFTNet respectively, GridNet accuracy dropped rapidly to below $50\%$ while FFTNet maintained near constant performance. Under conservative assumptions of errors associated with the beamsplitter ($\sigma_{BS} > 1\%$) and phaseshifter ($\sigma_{PS} > 0.01~\text{rad}$), a more robust network (FFTNet) can be favorable over one with greater expressivity (GridNet).

We then demonstrate, in Sec. \ref{sec:stack_trunc}, through modified unitary multipliers, TruncGrid and StackedFFT, that controlling for MZI layer depth, FFT-like designs are inherently more robust than grid-like ones.

To gain a better understanding of GridNet's sensitivity to imprecision, in Sec. \ref{sec:pos_sen}, we probed the response of the network to localized imprecisions by introducing error to small groups of MZIs at various locations. The sensitivity to imprecisions was found to be less affected by the MZIs' physical position within the grid and more so by the flow of the optical signal. We then demonstrated that beyond architectural designs, small procedural changes to the configuration of an ONN, such as shuffling the singular values, can change affect the its robustness. 

Our results, presented in this paper, provide clear guidelines for the architectural design of efficient, fault-resistant ONNs. In looking forward, it would be important to investigate algorithmic and training strategies as well. A central problem in deep learning is to design neural networks complex enough to model the data while being regularized to prevent over-fitting of noise in the training set\cite{goodfellow2016deep}. To this end, a wide variety of regularization techniques such as Dropout\cite{srivastava2014dropout}, Dropconnect\cite{wan2013regularization}, data augmentation, etc. have been developed. This problem parallels the trade-off between an ONN's expressivity and its robustness to imprecisions presented here. Indeed, an important conclusion in Sec. \ref{sec:pos_sen} is that in addition to architecture, even minor changes in the configuration of ONNs also have a great effect on the network's robustness to faulty components. 

The robustness of neural networks to perturbations \cite{fawzi2016robustness} is a well studied and open problem that is outside of the scope of this article on architectural design. Nevertheless, a complete analysis of ONNs with imprecise components requires an understanding of robustness due to architectural design as well as due to software training, possibly under a unifying framework. A natural direction for further exploration is to consider analogies to regularization in the context of imprecise photonic components and to focus on the development of algorithms and training strategies for error-resistant optical neural networks.

\section{Funding}
MRD was partially supported by the U. S. Army Research Laboratory and the U. S. Army Research Office under contract W911NF-13-1-0390.

\section{Code repository and results}
The code repository, results and scripts used to generate figures in this paper are freely available at \texttt{https://github.com/mike-fang/imprecise\_optical\_neural\_network}

\newpage
\bibliographystyle{unsrt}
\bibliography{refs}

\begin{thebibliography}{10}

\bibitem{farhat1985optical}
Nabil~H Farhat, Demetri Psaltis, Aluizio Prata, and Eung Paek.
\newblock Optical implementation of the hopfield model.
\newblock {\em Applied optics}, 24(10):1469--1475, 1985.

\bibitem{paquot2012optoelectronic}
Yvan Paquot, Francois Duport, Antoneo Smerieri, Joni Dambre, Benjamin
  Schrauwen, Marc Haelterman, and Serge Massar.
\newblock Optoelectronic reservoir computing.
\newblock {\em Scientific reports}, 2:287, 2012.

\bibitem{appeltant2011information}
Lennert Appeltant, Miguel~Cornelles Soriano, Guy Van~der Sande, Jan Danckaert,
  Serge Massar, Joni Dambre, Benjamin Schrauwen, Claudio~R Mirasso, and Ingo
  Fischer.
\newblock Information processing using a single dynamical node as complex
  system.
\newblock {\em Nature communications}, 2:468, 2011.

\bibitem{tait2017neuromorphic}
Alexander~N Tait, Thomas~Ferreira Lima, Ellen Zhou, Allie~X Wu, Mitchell~A
  Nahmias, Bhavin~J Shastri, and Paul~R Prucnal.
\newblock Neuromorphic photonic networks using silicon photonic weight banks.
\newblock {\em Scientific Reports}, 7(1):7430, 2017.

\bibitem{tait2014broadcast}
Alexander~N Tait, Mitchell~A Nahmias, Bhavin~J Shastri, and Paul~R Prucnal.
\newblock Broadcast and weight: an integrated network for scalable photonic
  spike processing.
\newblock {\em Journal of Lightwave Technology}, 32(21):3427--3439, 2014.

\bibitem{chang2018hybrid}
Julie Chang, Vincent Sitzmann, Xiong Dun, Wolfgang Heidrich, and Gordon
  Wetzstein.
\newblock Hybrid optical-electronic convolutional neural networks with
  optimized diffractive optics for image classification.
\newblock {\em Scientific reports}, 8(1):12324, 2018.

\bibitem{shen2017deep}
Yichen Shen, Nicholas~C Harris, Scott Skirlo, Mihika Prabhu, Tom Baehr-Jones,
  Michael Hochberg, Xin Sun, Shijie Zhao, Hugo Larochelle, and
  Marin~Solja\v{s}\'{c} Englund, Dirk.
\newblock Deep learning with coherent nanophotonic circuits.
\newblock {\em Nature Photonics}, 11(7):441, 2017.

\bibitem{goodfellow2016deep}
Ian Goodfellow, Yoshua Bengio, Aaron Courville, and Yoshua Bengio.
\newblock {\em Deep learning}, volume~1.
\newblock MIT Cambridge, 2016.

\bibitem{reck1994experimental}
Michael Reck, Anton Zeilinger, Herbert~J Bernstein, and Philip Bertani.
\newblock Experimental realization of any discrete unitary operator.
\newblock {\em Physical review letters}, 73(1):58, 1994.

\bibitem{clements2016optimal}
William~R Clements, Peter~C Humphreys, Benjamin~J Metcalf, W~Steven Kolthammer,
  and Ian~A Walmsley.
\newblock Optimal design for universal multiport interferometers.
\newblock {\em Optica}, 3(12):1460--1465, 2016.

\bibitem{barak2007quantum}
Ronen Barak and Yacob Ben-Aryeh.
\newblock Quantum fast fourier transform and quantum computation by linear
  optics.
\newblock {\em JOSA B}, 24(2):231--240, 2007.

\bibitem{carolan2015universal}
Jacques Carolan, Christopher Harrold, Chris Sparrow, Enrique
  Mart{\'\i}n-L{\'o}pez, Nicholas~J Russell, Joshua~W Silverstone, Peter~J
  Shadbolt, Nobuyuki Matsuda, Manabu Oguma, and Graham D. Marshall Mark G.
  Thompson Jonathan C F Matthews Toshikazu Hashimoto Jeremy L. O'Brien
  Anthony~Laing Itoh, Mikitaka.
\newblock Universal linear optics.
\newblock {\em Science}, 349(6249):711--716, 2015.

\bibitem{harris2017quantum}
Nicholas~C Harris, Gregory~R Steinbrecher, Mihika Prabhu, Yoav Lahini, Jacob
  Mower, Darius Bunandar, Changchen Chen, Franco~NC Wong, Tom Baehr-Jones,
  Michael Hochberg, Seth Lloyd, and Dirk Englund.
\newblock Quantum transport simulations in a programmable nanophotonic
  processor.
\newblock {\em Nature Photonics}, 11(7):447, 2017.

\bibitem{pai2018matrix}
Sunil Pai, Ben Bartlett, Olav Solgaard, and David~AB Miller.
\newblock Matrix optimization on universal unitary photonic devices.
\newblock {\em arXiv preprint arXiv:1808.00458}, 2018.

\bibitem{russell2017direct}
Nicholas~J Russell, Levon Chakhmakhchyan, Jeremy~L O'Brien, and Anthony Laing.
\newblock Direct dialling of haar random unitary matrices.
\newblock {\em New Journal of Physics}, 19(3):033007, 2017.

\bibitem{burgwal2017using}
Roel Burgwal, William~R Clements, Devin~H Smith, James~C Gates, W~Steven
  Kolthammer, Jelmer~J Renema, and Ian~A Walmsley.
\newblock Using an imperfect photonic network to implement random unitaries.
\newblock {\em Optics Express}, 25(23):28236--28245, 2017.

\bibitem{miller2015perfect}
David~AB Miller.
\newblock Perfect optics with imperfect components.
\newblock {\em Optica}, 2(8):747--750, 2015.

\bibitem{wilkes201660}
Callum~M Wilkes, Xiaogang Qiang, Jianwei Wang, Raffaele Santagati, Stefano
  Paesani, Xiaoqi Zhou, David~AB Miller, Graham~D Marshall, Mark~G Thompson,
  and Jeremy~L O’Brien.
\newblock 60 db high-extinction auto-configured mach--zehnder interferometer.
\newblock {\em Optics letters}, 41(22):5318--5321, 2016.

\bibitem{Fang2019Code}
Micael Y.-S. Fang.
\newblock Imprecise optical neural networks.
\newblock \url{https://github.com/mike-fang/imprecise_optical_neural_network},
  Mar 2019.

\bibitem{cooley1965algorithm}
James~W Cooley and John~W Tukey.
\newblock An algorithm for the machine calculation of complex fourier series.
\newblock {\em Mathematics of computation}, 19(90):297--301, 1965.

\bibitem{ma2013ultralow}
Yangjin Ma, Yi~Zhang, Shuyu Yang, Ari Novack, Ran Ding, Andy Eu-Jin Lim,
  Guo-Qiang Lo, Tom Baehr-Jones, and Michael Hochberg.
\newblock Ultralow loss single layer submicron silicon waveguide crossing for
  soi optical interconnect.
\newblock {\em Optics express}, 21(24):29374--29382, 2013.

\bibitem{gattass2008femtosecond}
Rafael~R Gattass and Eric Mazur.
\newblock Femtosecond laser micromachining in transparent materials.
\newblock {\em Nature photonics}, 2(4):219, 2008.

\bibitem{panusa2019photoinitiator}
Giulia Panusa, Ye~Pu, Jieping Wang, Christophe Moser, and Demetri Psaltis.
\newblock Photoinitiator-free multi-photon fabrication of compact optical
  waveguides in polydimethylsiloxane.
\newblock {\em Optical Materials Express}, 9(1):128--138, 2019.

\bibitem{connelly2007semiconductor}
Michael~J Connelly.
\newblock {\em Semiconductor optical amplifiers}.
\newblock Springer Science \& Business Media, 2007.

\bibitem{miller2017silicon}
David~AB Miller.
\newblock Silicon photonics: Meshing optics with applications.
\newblock {\em Nature Photonics}, 11(7):403, 2017.

\bibitem{bao2011monolayer}
Qiaoliang Bao, Han Zhang, Zhenhua Ni, Yu~Wang, Lakshminarayana Polavarapu,
  Zexiang Shen, Qing-Hua Xu, Dingyuan Tang, and Kian~Ping Loh.
\newblock Monolayer graphene as a saturable absorber in a mode-locked laser.
\newblock {\em Nano Research}, 4(3):297--307, 2011.

\bibitem{nair2010rectified}
Vinod Nair and Geoffrey~E Hinton.
\newblock Rectified linear units improve restricted boltzmann machines.
\newblock In {\em Proceedings of the 27th international conference on machine
  learning (ICML-10)}, pages 807--814, 2010.

\bibitem{arjovsky2016unitary}
Martin Arjovsky, Amar Shah, and Yoshua Bengio.
\newblock Unitary evolution recurrent neural networks.
\newblock In {\em International Conference on Machine Learning}, pages
  1120--1128, 2016.

\bibitem{xu2006optical}
Qianfan Xu and Michal Lipson.
\newblock Optical bistability based on the carrier dispersion effect in soi
  ring resonators.
\newblock In {\em Integrated Photonics Research and Applications}, page IMD2.
  Optical Society of America, 2006.

\bibitem{jiang2016analog}
Yunshan Jiang, Peter~TS DeVore, and Bahram Jalali.
\newblock Analog optical computing primitives in silicon photonics.
\newblock {\em Optics letters}, 41(6):1273--1276, 2016.

\bibitem{babaeian2018nonlinear}
Masoud Babaeian, Pierre-A Blanche, Robert~A Norwood, Tommi Kaplas, Patrick
  Keiffer, Yuri Svirko, Taylor~G Allen, Vincent~W Chen, San-Hui Chi, Joseph~W
  Perry, et~al.
\newblock Nonlinear optical components for all-optical probabilistic graphical
  model.
\newblock {\em Nature communications}, 9(1):2128, 2018.

\bibitem{lecun1998mnist}
Yann LeCun.
\newblock The mnist database of handwritten digits.
\newblock {\em http://yann.lecun.com/exdb/mnist/}.

\bibitem{paszke2017automatic}
Adam Paszke, Sam Gross, Soumith Chintala, Gregory Chanan, Edward Yang, Zachary
  DeVito, Zeming Lin, Alban Desmaison, Luca Antiga, and Adam Lerer.
\newblock Automatic differentiation in pytorch.
\newblock In {\em NIPS-W}, 2017.

\bibitem{dugas2001incorporating}
Charles Dugas, Yoshua Bengio, Fran{\c{c}}ois B{\'e}lisle, Claude Nadeau, and
  Ren{\'e} Garcia.
\newblock Incorporating second-order functional knowledge for better option
  pricing.
\newblock In {\em Advances in neural information processing systems}, pages
  472--478, 2001.

\bibitem{Cover:2006:EIT:1146355}
Thomas~M. Cover and Joy~A. Thomas.
\newblock {\em Elements of Information Theory (Wiley Series in
  Telecommunications and Signal Processing)}.
\newblock Wiley-Interscience, New York, NY, USA, 2006.

\bibitem{jing2017tunable}
Li~Jing, Yichen Shen, Tena Dubcek, John Peurifoy, Scott Skirlo, Yann LeCun, Max
  Tegmark, and Marin Solja{\v{c}}i{\'c}.
\newblock Tunable efficient unitary neural networks (eunn) and their
  application to rnns.
\newblock In {\em Proceedings of the 34th International Conference on Machine
  Learning-Volume 70}, pages 1733--1741. JMLR. org, 2017.

\bibitem{trabelsi2017deep}
Chiheb Trabelsi, Olexa Bilaniuk, Ying Zhang, Dmitriy Serdyuk, Sandeep
  Subramanian, Jo{\~a}o~Felipe Santos, Soroush Mehri, Negar Rostamzadeh, Yoshua
  Bengio, and Christopher~J Pal.
\newblock Deep complex networks.
\newblock {\em arXiv preprint arXiv:1705.09792}, 2017.

\bibitem{robbins1985stochastic}
Herbert Robbins and Sutton Monro.
\newblock A stochastic approximation method.
\newblock In {\em Herbert Robbins Selected Papers}, pages 102--109. Springer,
  1985.

\bibitem{simard2003best}
Patrice~Y Simard, Dave Steinkraus, and John~C Platt.
\newblock Best practices for convolutional neural networks applied to visual
  document analysis.
\newblock In {\em null}, page 958. IEEE, 2003.

\bibitem{srivastava2014dropout}
Nitish Srivastava, Geoffrey Hinton, Alex Krizhevsky, Ilya Sutskever, and Ruslan
  Salakhutdinov.
\newblock Dropout: a simple way to prevent neural networks from overfitting.
\newblock {\em The Journal of Machine Learning Research}, 15(1):1929--1958,
  2014.

\bibitem{flamini2017benchmarking}
Fulvio Flamini, Nicol{\`o} Spagnolo, Niko Viggianiello, Andrea Crespi, Roberto
  Osellame, and Fabio Sciarrino.
\newblock Benchmarking integrated linear-optical architectures for quantum
  information processing.
\newblock {\em Scientific Reports}, 7(1):15133, 2017.

\bibitem{flamini2015thermally}
Fulvio Flamini, Lorenzo Magrini, Adil~S Rab, Nicol{\`o} Spagnolo, Vincenzo
  D'ambrosio, Paolo Mataloni, Fabio Sciarrino, Tommaso Zandrini, Andrea Crespi,
  Roberta Ramponi, and Roberto Osellame.
\newblock Thermally reconfigurable quantum photonic circuits at telecom
  wavelength by femtosecond laser micromachining.
\newblock {\em Light: Science \& Applications}, 4(11):e354, 2015.

\bibitem{walls2007quantum}
Daniel~F Walls and Gerard~J Milburn.
\newblock {\em Quantum optics}.
\newblock Springer Science \& Business Media, 2007.

\bibitem{fawzi2016robustness}
Alhussein Fawzi, Seyed-Mohsen Moosavi-Dezfooli, and Pascal Frossard.
\newblock Robustness of classifiers: from adversarial to random noise.
\newblock In {\em Advances in Neural Information Processing Systems}, pages
  1632--1640, 2016.

\bibitem{wan2013regularization}
Li~Wan, Matthew Zeiler, Sixin Zhang, Yann Le~Cun, and Rob Fergus.
\newblock Regularization of neural networks using dropconnect.
\newblock In {\em International Conference on Machine Learning}, pages
  1058--1066, 2013.

\bibitem{kikuchi2012characterization}
Kazuro Kikuchi.
\newblock Characterization of semiconductor-laser phase noise and estimation of
  bit-error rate performance with low-speed offline digital coherent receivers.
\newblock {\em Optics Express}, 20(5):5291--5302, 2012.

\bibitem{larson2013narrow}
MC~Larson, Yan Feng, Ping-Chiek Koh, Xiao-dong Huang, Michael Moewe, Alex
  Semakov, Aditi Patwardhan, Eddie Chiu, Ashish Bhardwaj, Kit Chan, et~al.
\newblock Narrow linewidth high power thermally tuned sampled-grating
  distributed bragg reflector laser.
\newblock In {\em 2013 Optical Fiber Communication Conference and Exposition
  and the National Fiber Optic Engineers Conference (OFC/NFOEC)}, pages 1--3.
  IEEE, 2013.

\bibitem{selden1967pulse}
AC~Selden.
\newblock Pulse transmission through a saturable absorber.
\newblock {\em British Journal of Applied Physics}, 18(6):743, 1967.

\bibitem{hubara2017quantized}
Itay Hubara, Matthieu Courbariaux, Daniel Soudry, Ran El-Yaniv, and Yoshua
  Bengio.
\newblock Quantized neural networks: Training neural networks with low
  precision weights and activations.
\newblock {\em The Journal of Machine Learning Research}, 18(1):6869--6898,
  2017.

\bibitem{rastegari2016xnor}
Mohammad Rastegari, Vicente Ordonez, Joseph Redmon, and Ali Farhadi.
\newblock Xnor-net: Imagenet classification using binary convolutional neural
  networks.
\newblock In {\em European Conference on Computer Vision}, pages 525--542.
  Springer, 2016.

\end{thebibliography}

\section*{Appendix}
\begin{appendices}
\section{MZI transfer matrix}
\label{app:transfer_matrix}
Because MZIs are comprised of beampslitters and phaseshifters, we state their respective transfer matrix first.
\begin{align}
    U_{BS}(r) =
    \begin{pmatrix}
    r & i t\\
    it & r
    \end{pmatrix}
    \label{eq:U_BS}
\end{align}
where $t \equiv \sqrt{1-r^2}$
and
\begin{align}
    U_{PS}(\theta) =
    \begin{pmatrix}
    e^{i\theta} & 0\\
    0 & 1
    \end{pmatrix}.
\end{align}
With the construction of PS-BS-PS-BS (Fig. \ref{fig:grid_arch}, inset), the MZI transfer matrix is the following matrix product:
\begin{align}
    U_{MZI}(\theta, \phi; r, r') &= U_{BS}(r) U_{PS}(\theta) U_{BS}(r') U_{PS}(\phi)\\
    &= \begin{pmatrix}
    r & i t\\
    it & r
    \end{pmatrix}
    \begin{pmatrix}
    e^{i\theta} & 0\\
    0 & 1
    \end{pmatrix}
    \begin{pmatrix}
    r' & i t'\\
    it' & r'
    \end{pmatrix}
    \begin{pmatrix}
    e^{i\phi} & 0\\
    0 & 1
    \end{pmatrix}\\
    &=
    \begin{pmatrix}
     e^{i \phi } \left(e^{i \theta } r r'-t t'\right) & i \left(t \rho +e^{i \theta } r t'\right) \\
     i e^{i \phi } \left(e^{i \theta } t r'+r t'\right) & r \rho -e^{i \theta } t t' \\
    \end{pmatrix}
    \label{eq:U_MZI_full}
\end{align}
Assuming that the beamsplitter ratios are 50:50, we can take $r=t=1/\sqrt2$ so that 
\begin{align}
    U_{BS} \equiv U_{BS}\left(\pi/2\right) = 
    \frac{1}{\sqrt{2}}
    \begin{pmatrix}
   1 & i\\
   i & 1
    \end{pmatrix}
\end{align}
and therefore,
\begin{align}
    U_{MZI}(\theta, \phi) = 
    i e^{i\theta/2}
   \begin{pmatrix}
    e^{i\phi}\sin\frac\theta2 & \cos\frac\theta2\\
    e^{i\phi}\cos\frac\theta2 & -\sin\frac\theta2
    \end{pmatrix}
\end{align}
In our convention, the transmission and reflection coefficient is
\begin{align}
    T = |\cos \theta /2|^2 \text{~and~} R = |\sin \theta /2|^2
\end{align}
respectively. In particular, the MZI is in the bar state ($T = 0$) when $\theta = \pi$ and in the cross state ($T = 1$) when $\theta = 0$.

However, in other conventions, the beamsplitter is often taken to be the Hardamard gate.
\begin{align}
    H = \frac1{\sqrt{2}}
    \begin{pmatrix}
    1 & 1\\
    1 & -1
    \end{pmatrix}.
\end{align}
We note however, that
\begin{align}
    U_{BS} = 
    \begin{pmatrix}
    1 & 0 \\
    0 & i
    \end{pmatrix}
    H
    \begin{pmatrix}
    1 & 0 \\
    0 & i
    \end{pmatrix}
    = U_{PS}(-\pi/2) H U_{PS}(-\pi/2)
\end{align}
up to a global phase. We then can express the MZI transfer matrix as
\begin{align}
U_{MZ}(\theta, \phi) = U_{PS}(-\pi/2) H U_{PS}(\theta - \pi) H U_{PS}(\phi - \pi/2).
\end{align}
Note in this convention the internal phase shift is now $\theta + \pi$ and thus the bar and cross states are now at $\theta =0$ and $\theta =\pi$ respectively.

\section{Laser phase noise}
\label{app:laser_phase_noise}
The variance in phase for typical lasers can be modeled as\cite{kikuchi2012characterization}
\begin{align}
    \sigma_\phi(\tau)^2 = 2\pi \cdot \delta f \cdot \tau.
\end{align}
Here, $\tau$ is the time of integration and $\delta f$ the linewidth of the laser. For an order or magnitude calculation, we ignore the refractive index and take $\tau = L / c$ where $L$ is the distance between two subsequent phaseshifters on an MZI. Again, as an order of magnitude estimate, we take $L = 100\mu\text{m} = 10^{-4} \text{m}$ and thus $\tau \approx 3 \times 10^{-13}$. We wish to solve for the linewidth required for $\sigma_\phi = 0.01\text{rad}$:
\begin{align}
    \sigma_\phi^2 = 10^{-4} &= 2 \pi \cdot \delta f \cdot \tau \\
    &\approx 6 \cdot 3 \times 10^{-13} s \delta f\\
    \delta f &\approx 5 \times 10^{7} \text{Hz}\\
    &= 50 \text{MHz}.
\end{align}
A linewidth of 50 MHz is easily achieved by modern lasers. For example, Bragg reflector lasers have been shown to achieve a linewidth of 300 kHz \cite{larson2013narrow}. Thus, the contribution to phase noise from the laser is roughly two orders of magnitude smaller than that from MZIs.
%==================== Saturable Absorption ==================== 
\section{Approximating saturable absorption}
\begin{figure}[h!]
    \centering
    \includegraphics[width=0.8\columnwidth]{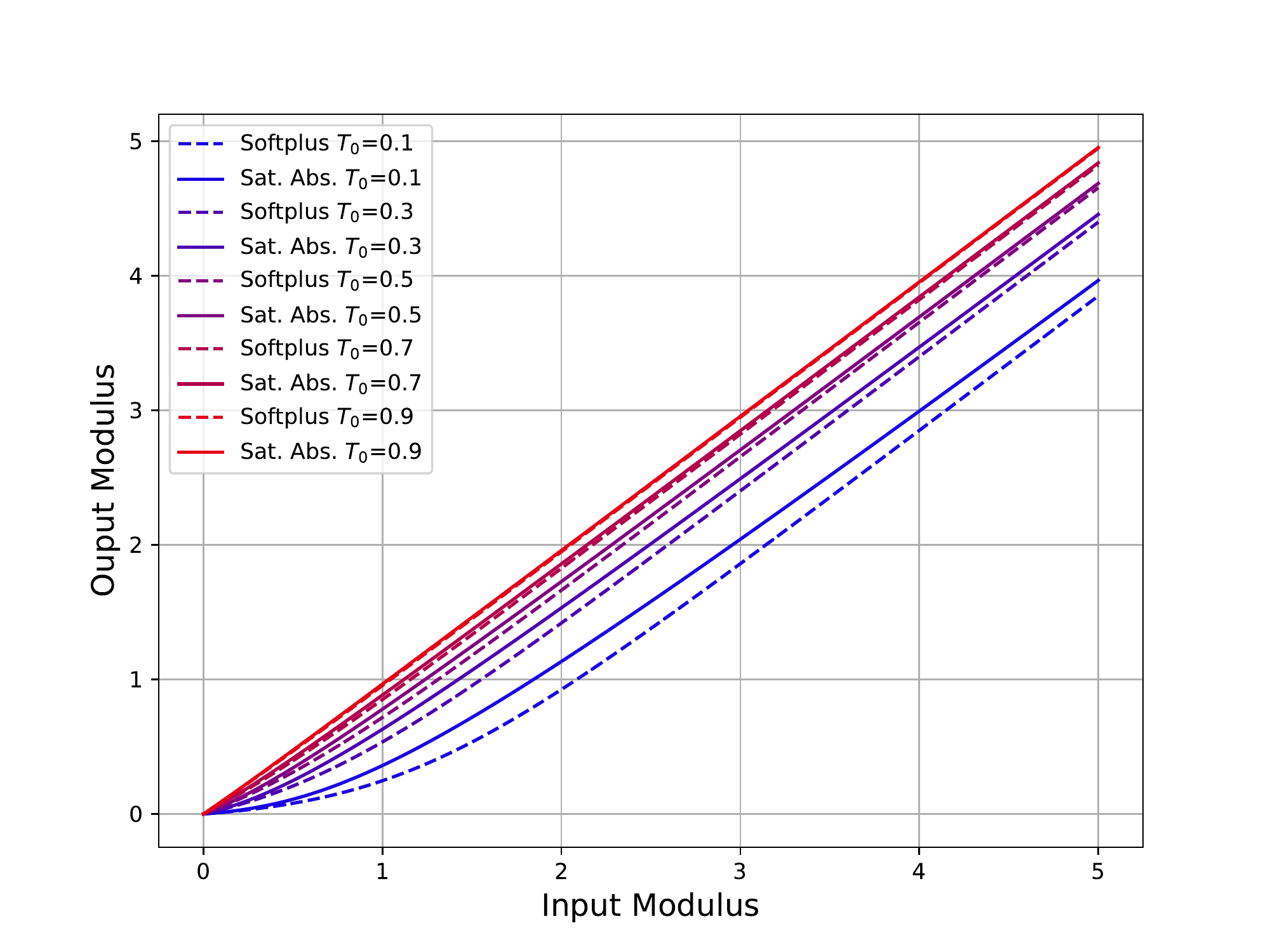}
    \caption{The saturable absorption response curve compared to the corresponding Softplus approximation with various values of $T$.}
    \label{fig:softplus_approx}
\end{figure}
\label{app:sat_abs}
Saturable absorption can be modeled by the relation\cite{selden1967pulse}
\begin{align}
    u_0 = \frac12 \frac{\log(T/T_0)}{1-T}
\end{align}
where $T = u/u_0$ and $u = \sigma \tau_s I$ and $u_0 = \sigma\tau_s I_0$. $I_0, I$ are the incidental and transmitted intensities, respectively. The above equation can be solved to be
\begin{align}
    u = \frac12 W (2T_0 u_0 e^{2u_0}) \equiv f(u_0)
\end{align}
Where $W$ is the product log function or Lambert W function. However, since $W$ is not readily available in most deep learning libraries and difficult to implement, we wish to approximate the above by the shifted and biased Softplus non-linearity of the form
\begin{align}
\sigma(u) = \beta^{-1} \log\left(\frac{1+e^{\beta(u-u_0)}}{1+e^{-\beta u_0}}\right).
\end{align}

The bias of $-\beta^{-1} \log(1+e^{-\beta u_0})$ was chosen to ensure that $\sigma(0) = f(0) = 0$. We now choose $\beta$ and $u_0$ to ensure that
\begin{enumerate}
    \item $\sigma'(0) = f'(0) = T_0$,
    \item $\lim_{u\rightarrow\infty}\sigma(u) - u = \lim_{u\rightarrow\infty}f(u) - u = \frac12 \log T_0$.
\end{enumerate}
The derivative of $\sigma(u)$ is easily found to be
\begin{align}
   \sigma'(0) &= \frac{e^{-\beta u_0}}{1+e^{-\beta u_0}}\\ 
    &= (1 + e^{\beta u_0})^{-1}.
\end{align}
Requiring that it equals to $f'(0) = T_0$ allows us to solve for
\begin{align}
    u_0 = \beta^{-1} \log\left(T_0^{-1} - 1\right).
    \label{eq:u_0}
\end{align}
Next, in the large $u$ limit, the biased Softplus converges to 
\begin{align}
    \sigma(u) \rightarrow (u-u_0) -\beta^{-1} \log(1 + e^{-\beta u_0}).
\end{align}
Solving for equality with $f(u) \rightarrow u + \frac12 \log T_0$ gives
\begin{align}
    u_0 + \beta^{-1}\log(1 + e^{-\beta u_0}) &= - \frac12 \log T_0\\
    \beta u_0 + \log\left(1+ \frac{1}{T_0^{-1} - 1}\right) &= -\frac12 \beta \log T_0\\
    \log(T_0) &= -\frac12 \beta \log T_0\\
    \beta &= 2.
\end{align}
Going back to Eq. \eqref{eq:u_0}, we obtain
\begin{align}
   u_0 = \frac12 \log(T_0^{-1} - 1).
\end{align}
Fig. \ref{fig:softplus_approx} plots the saturable absorption response curve compared to the Softplus approximation derived above.

\section{Confusion matrices}
\label{app:conf_mat}
\begin{figure}[ht!]
    \centering
    \subfloat[Ideal GridNet]{%
    \label{fig:cm_grid_0}%
    \includegraphics[height=1.2in]{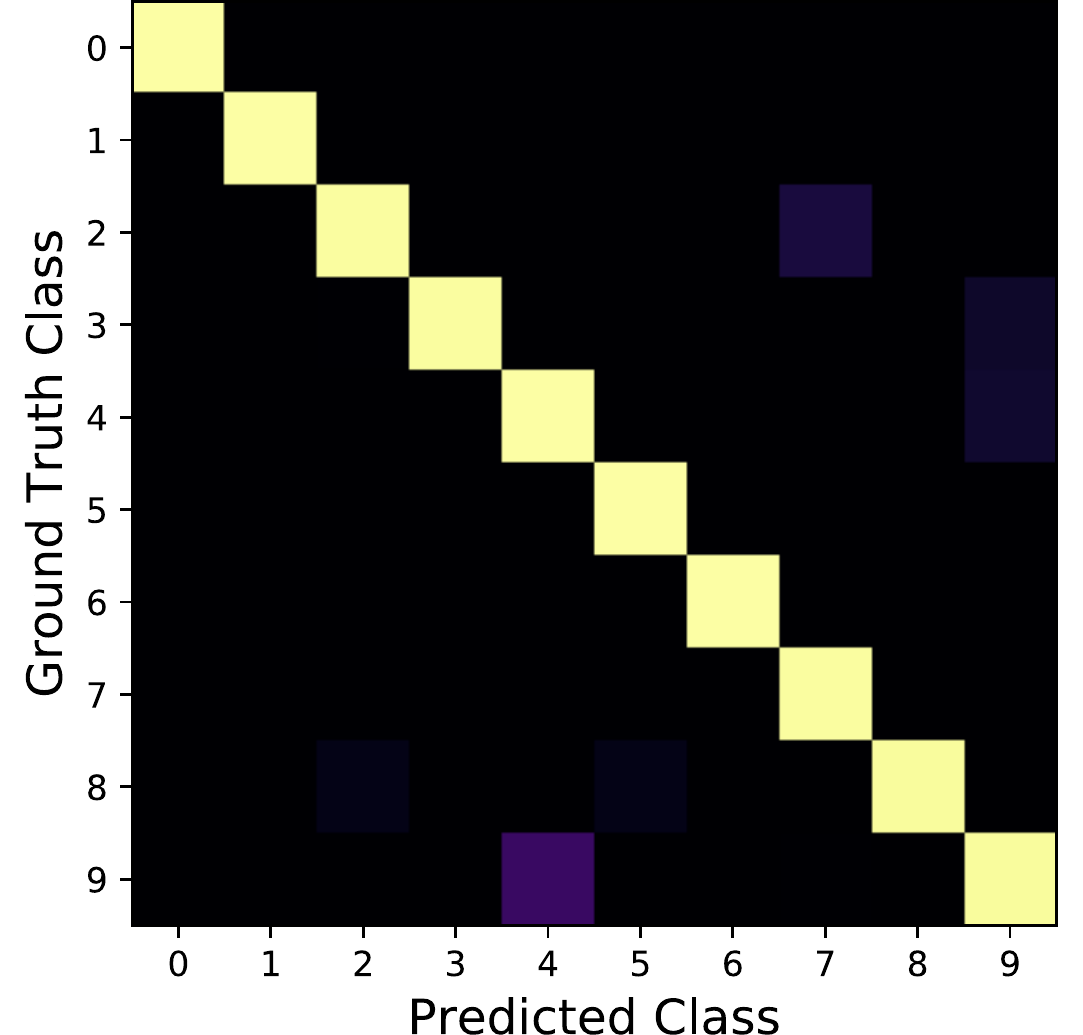}}%
    \subfloat[GridNet : $\sigma = 0.01$]{%
    \label{fig:cm_grid_1}%
    \includegraphics[height=1.2in]{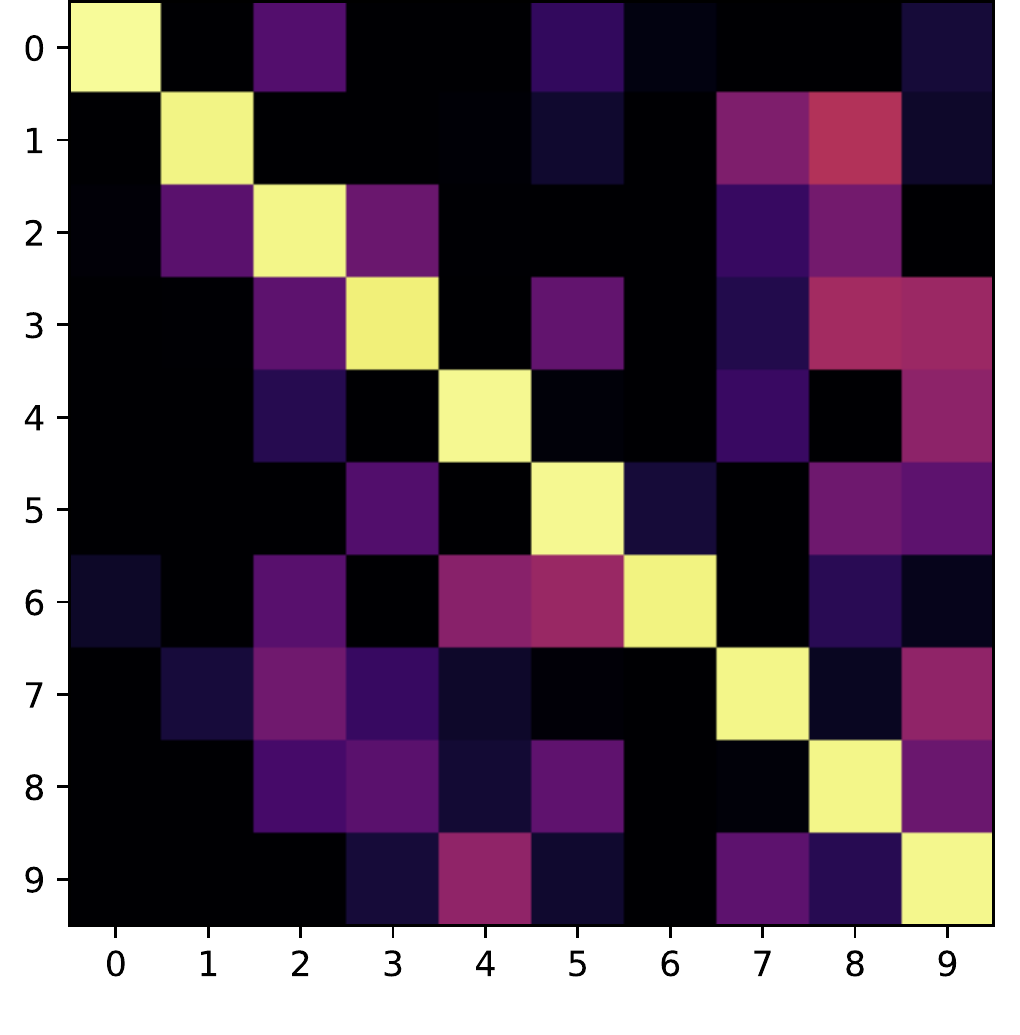}}%
    \subfloat[GridNet : $\sigma = 0.02$]{%
    \label{fig:cm_grid_2}%
    \includegraphics[height=1.2in]{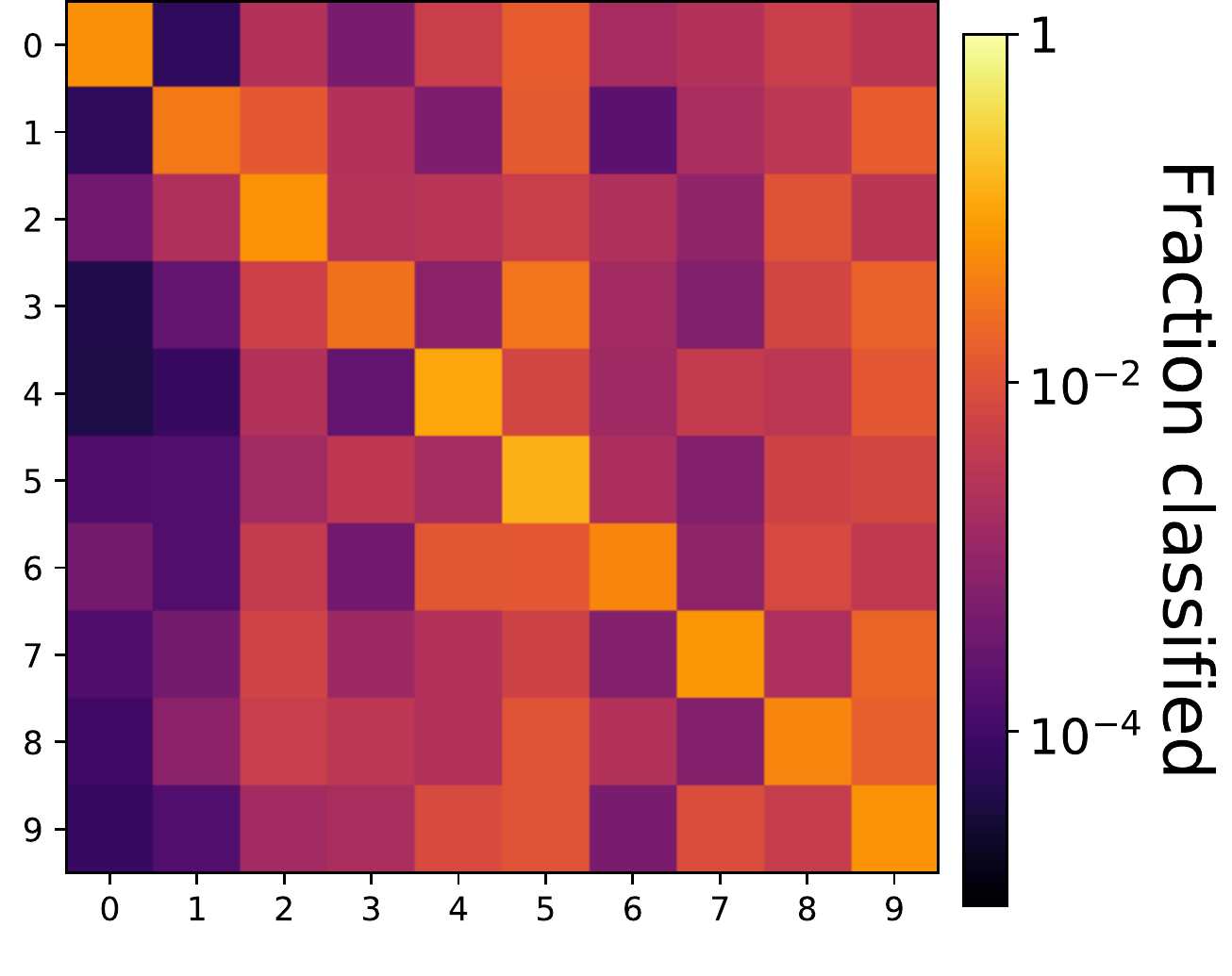}}%
    \\
    \subfloat[Ideal FFTNet]{%
    \label{fig:cm_grid_0}%
    \includegraphics[height=1.2in]{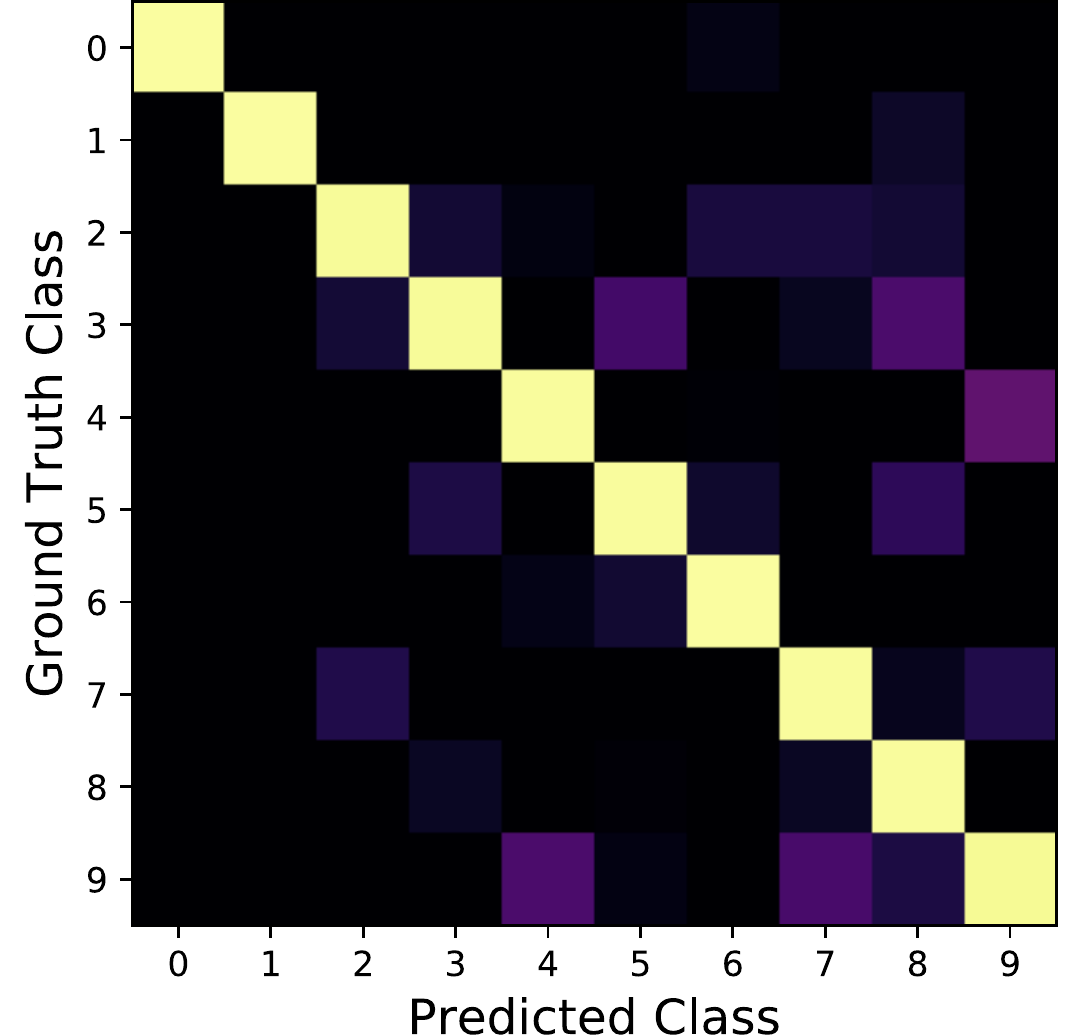}}%
    \subfloat[FFTNet : $\sigma = 0.01$]{%
    \label{fig:cm_fft_1}%
    \includegraphics[height=1.2in]{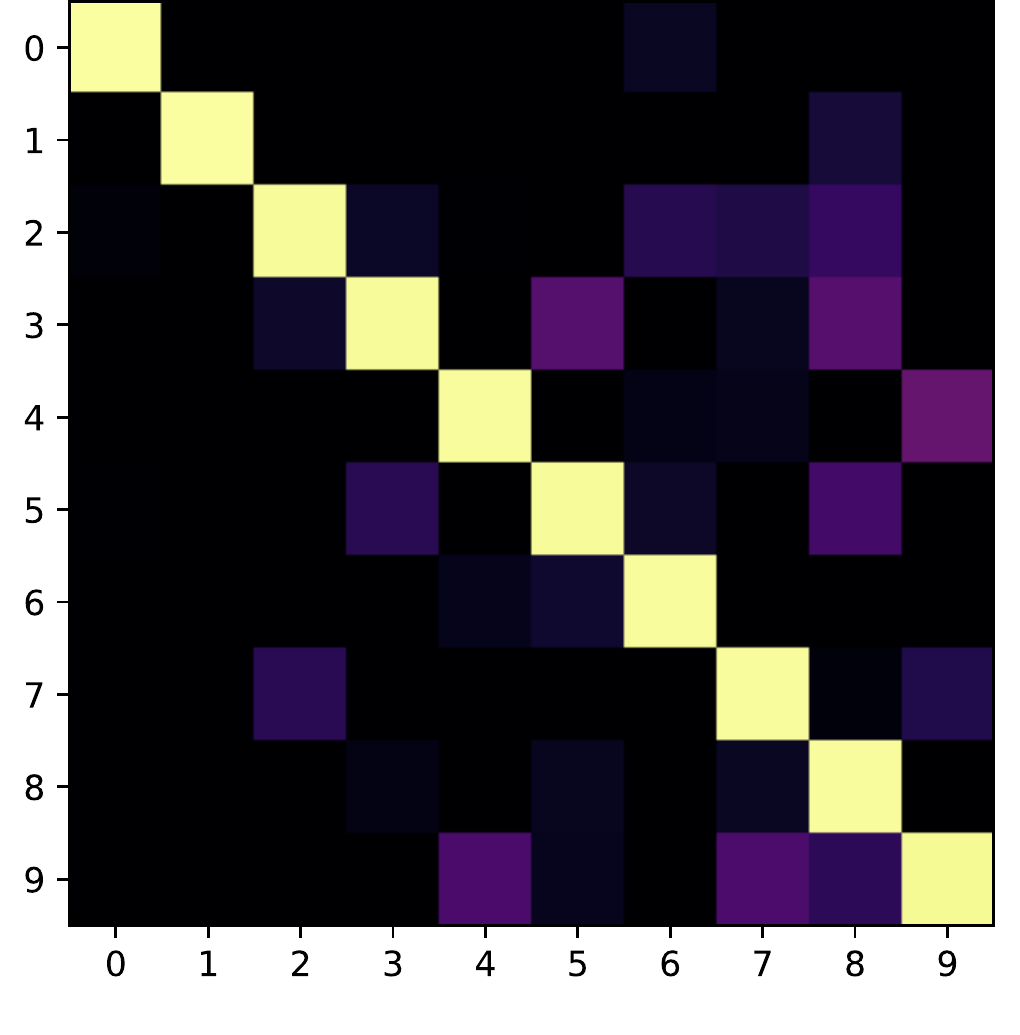}}%
    \subfloat[FFTNet : $\sigma = 0.02$]{%
    \label{fig:cm_fft_2}%
    \includegraphics[height=1.2in]{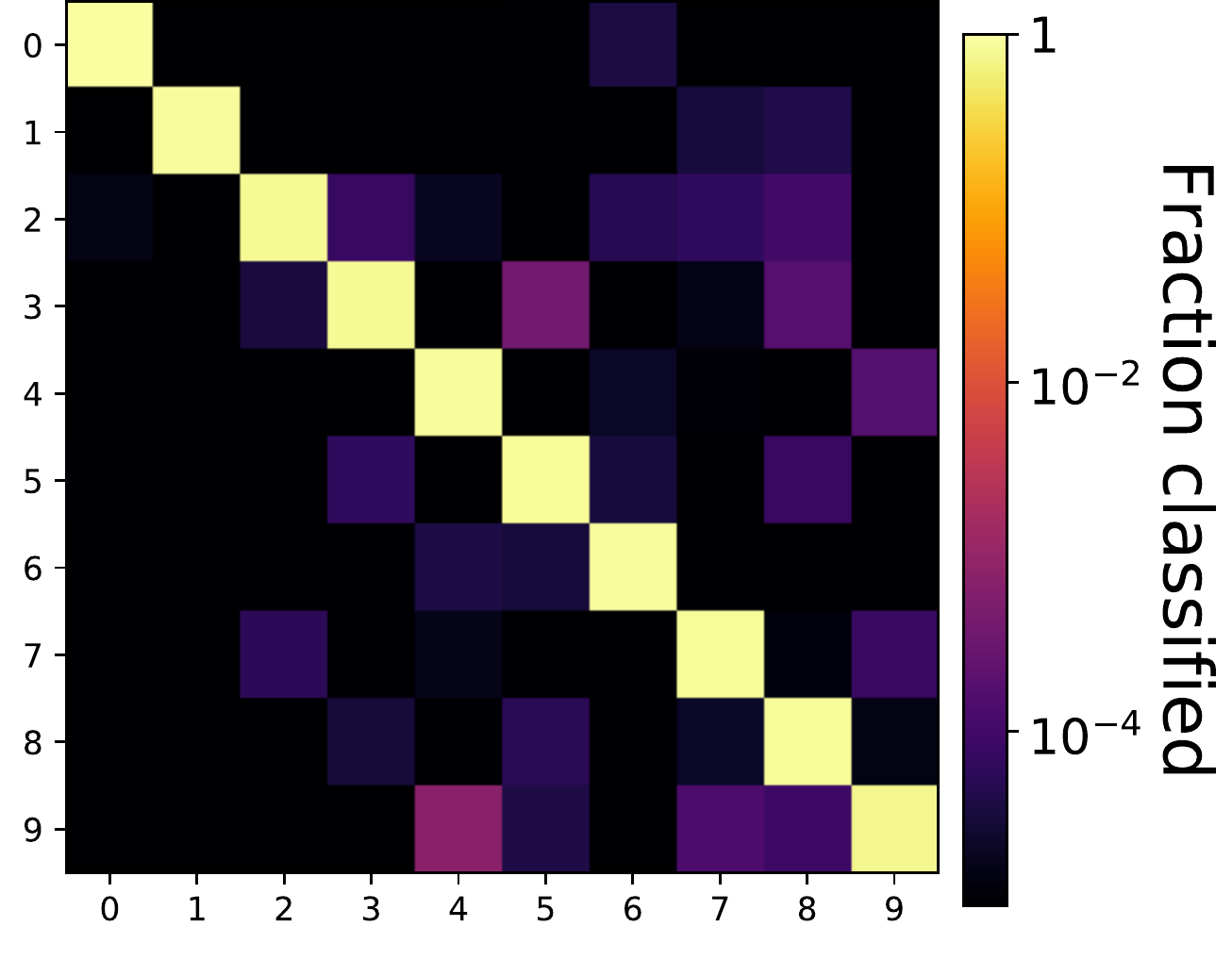}}%
    \caption{The degradation of ONN outputs visualized through confusion matrices. Each confusion matrix shows how often each target class (row) is predicted as each of the ten possible classs (column). Both networks, GridNet (a, b, c) and FFTNet (d, e, f) are evaluated. First in the ideal case (a, d) then, with increasing errors (b, e and c ,f). Note the logarithmic scaling.}
    \label{fig:conf_mat}
\end{figure}
To investigate the degradation of the networks due to imprecisions, we produce confusion matrices for both networks in the ideal case, with no imprecisions, and with different levels of error. $\sigma_{BS} = 1\%$, $\sigma_{PS} = 0.01\text{rad}$ and $\sigma_{BS} = 2\%$, $\sigma_{PS} = 0.02\text{rad}$ (Fig. \ref{fig:conf_mat}).

The imprecisions were simulated 10 times and the mean of the output was used in generating the confusion matrices.

\section{Quantization error}
\label{app:quant_err}

\begin{figure}
    \centering
    \includegraphics[width=.8\columnwidth]{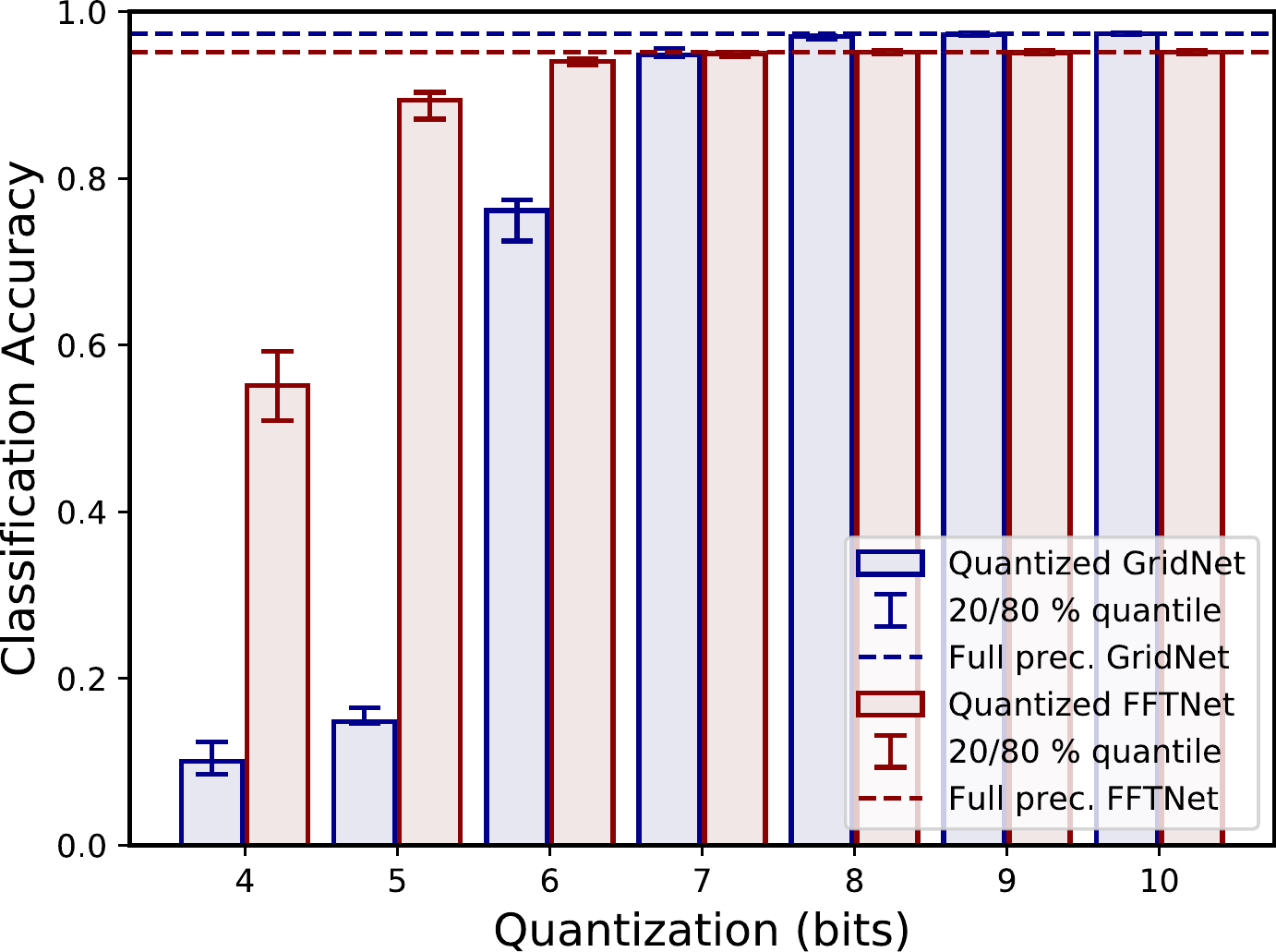}
    \caption{The effects of quantization is shown for both GridNet and FFTNet. 10 instances of GridNet (blue) and FFTNet (red) were trained then quantized to varying levels. The mean classification accuracy at each level is shown by bar plots. The 20-80\% quantiles are shown with error bars. The dotted horizontal line denotes the full precision accuracy.}
    \label{fig:quant_err}
\end{figure}
In this section, we explore the quantization error introduced by thermo-optic phaseshifters. Assuming a linear relationship between refractive index and temperature and quadratic relationship between temperature and voltage, we have
\begin{align*}
   \theta &\propto V^2\\
   \theta &= 2\pi \left(\frac{V}{V_{2\pi}} \right)^2 \\
   &\equiv 2\pi u^2\\
   \sqrt{\frac{\theta}{2\pi}} &= u.
\end{align*}
We have taken $V_{2\pi}$ to be the voltage required for a $2\pi$ phaseshift and defined the dimensionless voltage $u = V/V_{2\pi}$. Assuming that the voltage can be set with $B$-bit precisions, $u$ must take on values of
\begin{align*}
    u \in \{2^{-B} i : i = 0, \dots, 2^{B} - 1\}.
\end{align*}
The quantization procedure then takes
\begin{align*}
   \theta \rightarrow \tilde \theta \in \left\{\frac{2\pi}{2^{2B}} i^2 : i=0, \dots, 2^{B} - 1 \right\}.
\end{align*}

To evaluate the sensitivity to quantization, we quantized GridNet and FFTNet with varying levels of precision. Since quantization is deterministic, we trained 10 instances of both networks with randomized initialization and thus different configuration but similar ideal accuracies ($\sim 95\%$ and $\sim 98\%$). The networks were then quantized at varying levels -- from 4 to 10 bits. Their classification accuracy at each level is shown in Fig. \ref{fig:quant_err}.

Similar to results with simulated Gaussian noise, FFTNet is more robust than GridNet. Note that in this case, the quantization was applied after training has finished. Neural networks in which quantization happens as part of the training procedure has been demonstrated to have accuracies very near their full precision counterpart, down to even binary weights \cite{hubara2017quantized, rastegari2016xnor}.

\section{Empirical distribution of phases}
\label{app:phase_distr}

\begin{figure}[h!]
    \centering
    \subfloat[]{%
    \label{fig:grid_phase_pos}%
    \includegraphics[height = 1.5in]{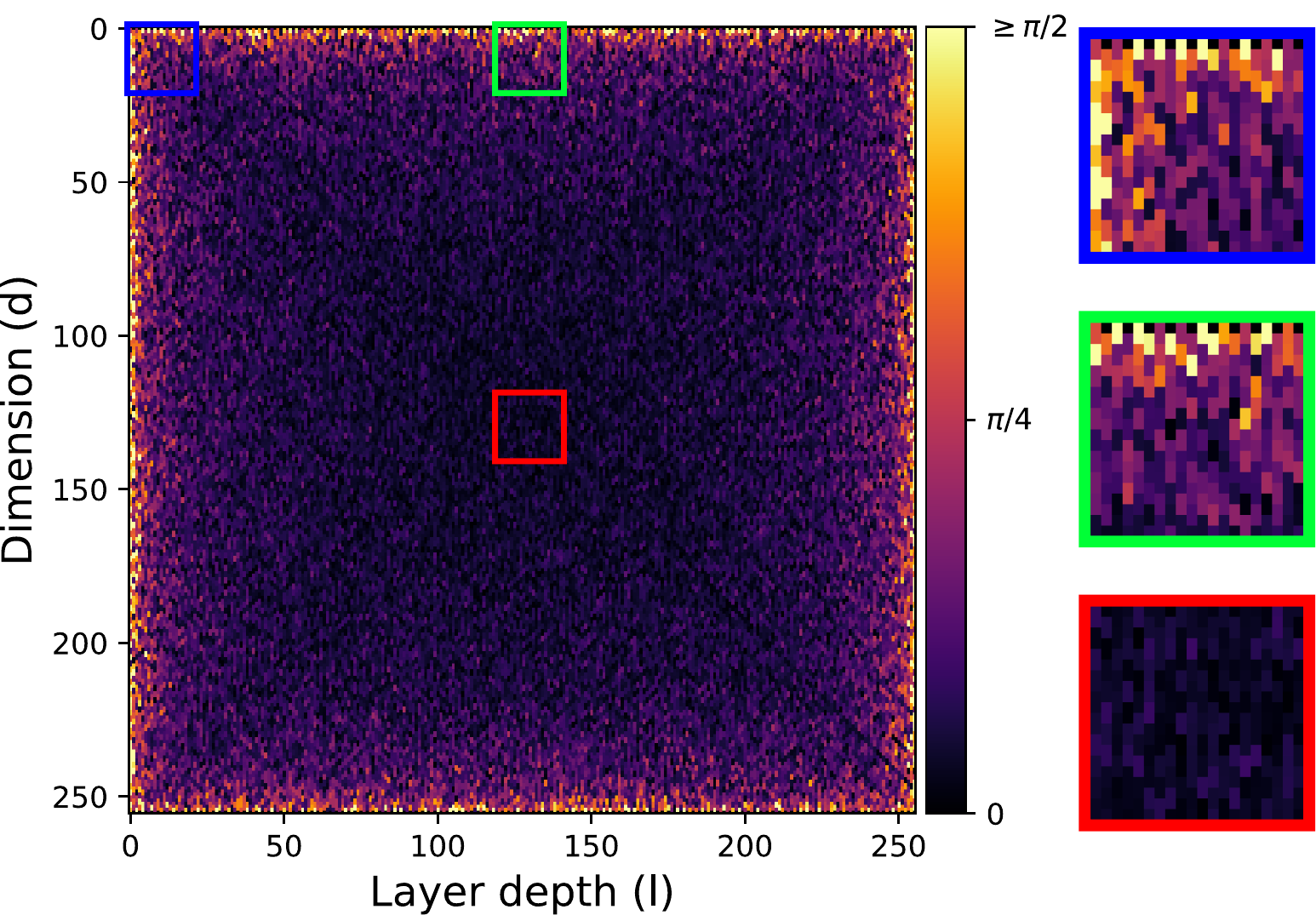}}%
    \subfloat[]{%
    \label{fig:grid_phase_hist}%
    \includegraphics[height = 1.5in]{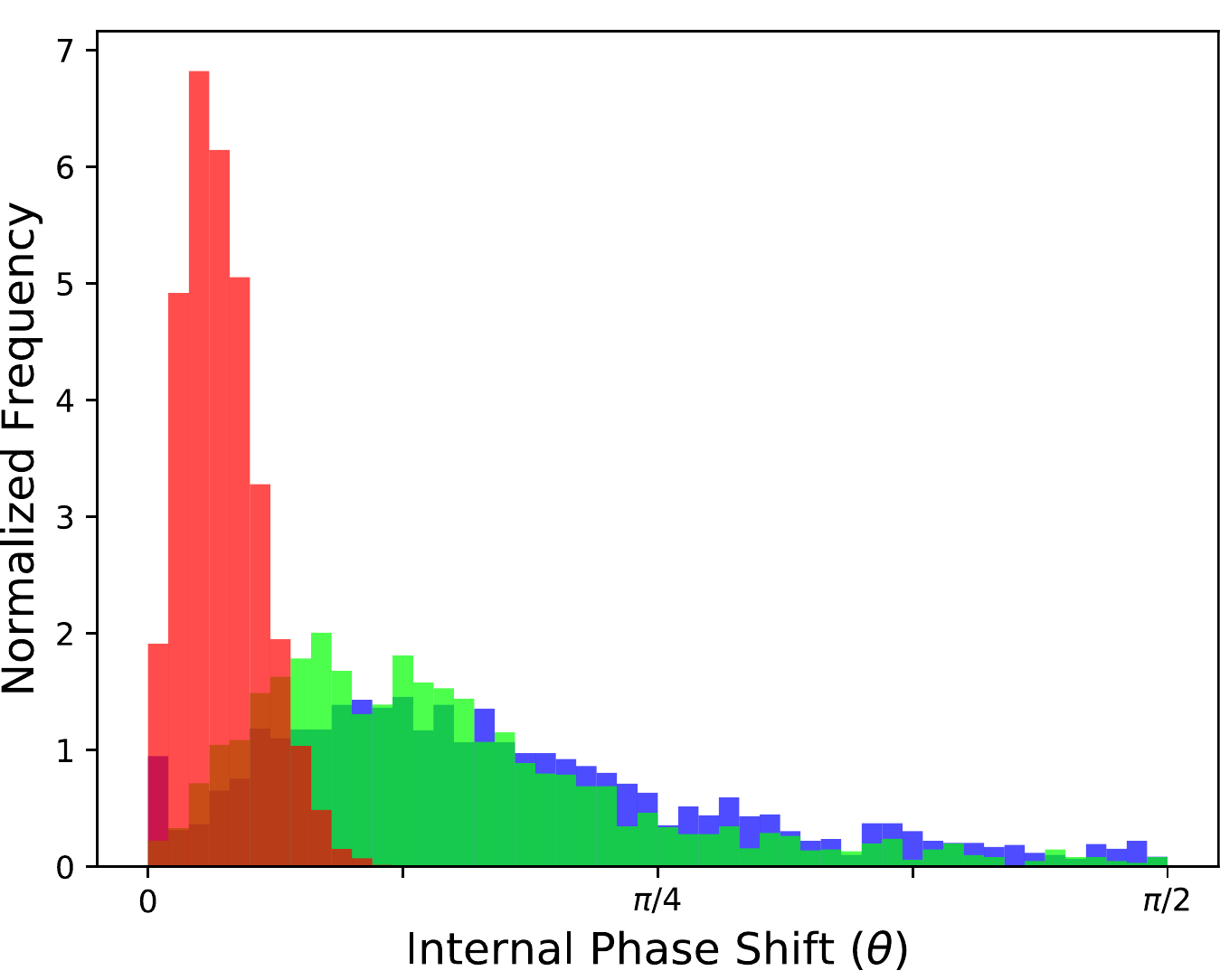}}%
    \caption{The central MZIs of GridNet has lower variance in internal phase shifts ($\theta$). a) The spatial distribution of internal phase shift ($\theta_{d, l}$) of MZIs in $U_2$ of GridNet. Reference Fig. \ref{fig:grid_arch} for coordinates and Fig. \ref{fig:network_arch} for location of $U_2$ in context of network architecture. b) Histogram of phase shifts near the center (red), edge (green), and corner (blue) of the GridUnitary multiplier. These phases are obtained from multiple instances of trained GridNets with random initialization.}
\end{figure}
\begin{figure}[h!]
    \centering
    \subfloat[]{%
    \label{fig:fft_phase_pos}%
    \includegraphics[height=1.5in]{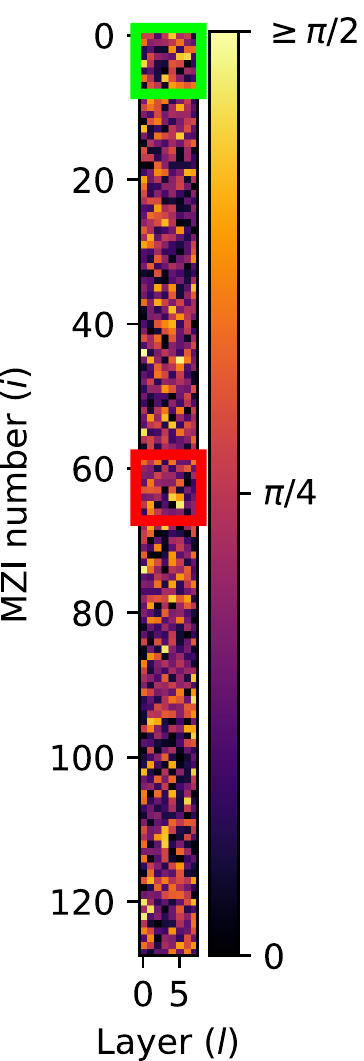}}%
    \subfloat[]{%
    \label{fig:fft_phase_hist}%
    \includegraphics[height=1.5in]{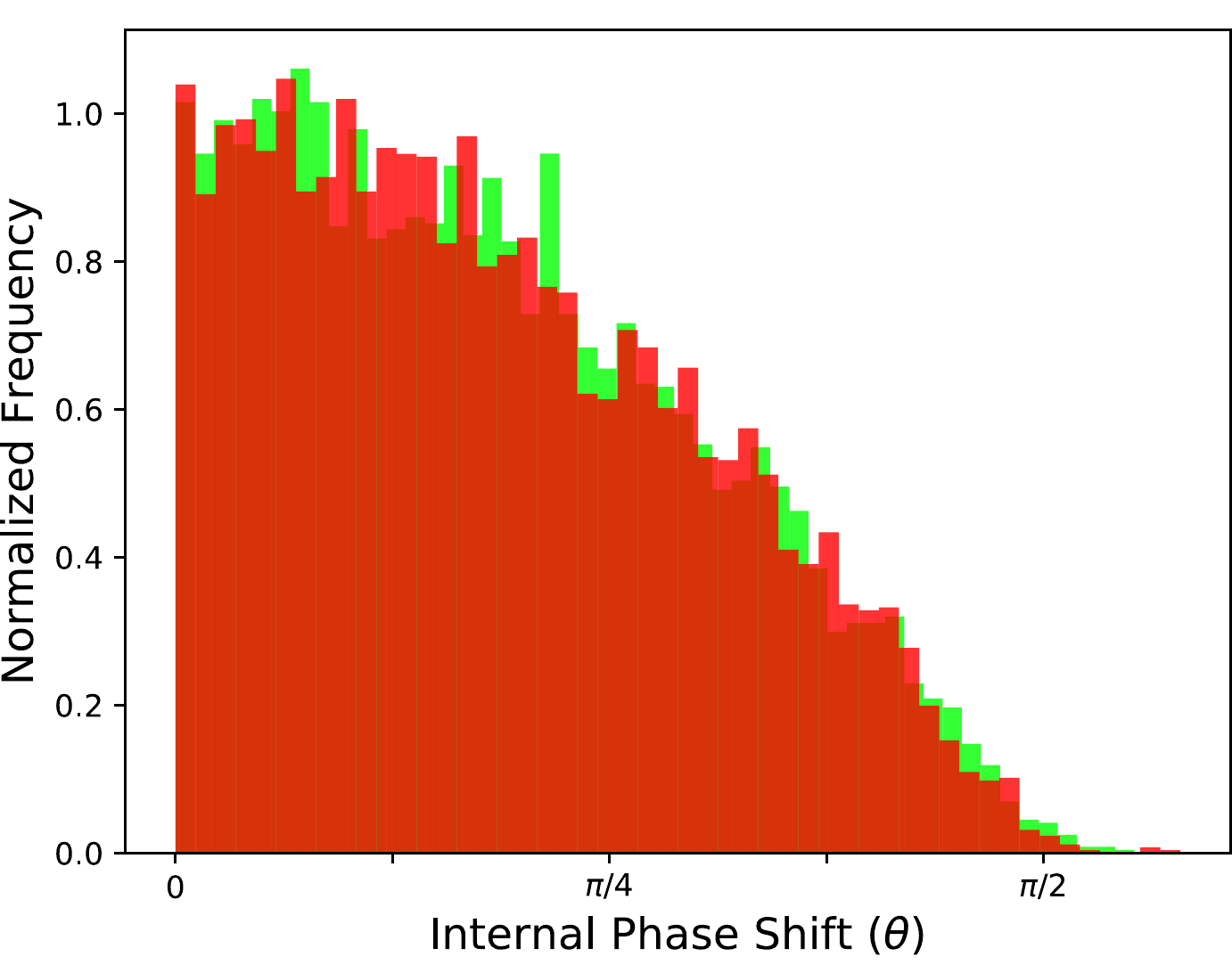}}%
    \caption{The variance of internal phase shifts of FFTNet is uniform spatially (a) Spatial distribution of phase shifts for a FFTUnitary multiplier. The MZIs are ordered as shown in Fig. \ref{fig:fft_arch}. (b) Histogram of phase shifts of FFTUnitary near the center (red) and top (green). These phases are obtained from mulitple trained FFTNets with random initialization.}
    \label{fig:fft_phase}
\end{figure}

Analyses has been done on the distribution of the internal phase shift ($\theta$) of MZIs of GridUnitary multipliers when used to implement randomly sampled unitary matrices \cite{russell2017direct, burgwal2017using, pai2018matrix}. It was shown that the phases are not uniformly distributed spatially. To be more concrete, We denote $d$ the waveguide number and $l$ the layer number (see Fig. \ref{fig:grid_arch}). The distribution of the MZI reflectivity ($r = \sin(\theta/2)$) is\cite{russell2017direct}
\begin{align}
r_{d, l} \sim \text{Beta}(1, \beta_{d, l}).
\label{eq:theory_distr}
\end{align}
For large dimensions $N$,
\begin{align}
\beta&\approx N - 2 \max(|d - N/2|, |l - N/2|)\\
&= N - 2 ||(d, l) - (N/2, N/2)||_{\infty}.
\end{align}
$\beta$ decreases from $N$ at the center of the grid layout to $0$ at the edge of the grid. For large $\beta$ (i.e. near the center), the mean and variance of $r_{d, l}$ are approximately
$$
\mu_r \approx \beta^{-1} ; \sigma^2_r \approx \beta^{-2}.
$$
Consequently, the reflectivity, and therefore the internal phases, of MZIs near the center of Gird Unitary multipliers are distributed very close to 0, with low variance. This effect is magnified with larger dimensions $N$. 

This result was derived with the assumption of Haar-random unitary matrices. Such a distribution is not guaranteed and not expected for layers of trained neural networks. (Fig. \ref{fig:grid_phase_pos}) shows the spatial distribution of phases in the GridUnitary multiplier $U_2$ (see Fig. \ref{fig:network_arch}). While the empirical histogram (Fig. \ref{fig:grid_phase_hist}) does not match the theoretical distribution (Eq. \eqref{eq:theory_distr}), the general trend of lower variance near the center of GridUnitary multipliers is evident. This is claimed to translates to a lower tolerance for error\cite{pai2018matrix}. 

A similar analysis was conducted for FFTNet. Immediately we notice that the distribution of phase shifts is mostly uniform across the MZIs (Fig. \ref{fig:fft_phase_pos}). This can be attributed to the non-local connectivity of FFTUnitary multipliers. Histograms constructed from an ensemble of 100 trained FFTNets with random initial weights (Fig. \ref{fig:fft_phase_hist}) confirms this observation. The histogram for the region near the center (red) is nearly identical to the top (green).

We reiterate the distinction, made in Section \ref{sec:pos_sen}, between pre-fabrication error tolerance and the sensitivity of error introduced post-fabrication. Pertinent to the first concept is how well the network can be optimized after a known set of imperfections are introduced to the network. The latter concept, which is relevant for our discussion, describes the sensitivity of the network with no further reconfiguration to unknown errors. In contrast to pre-fabrication error tolerance, our analysis in \ref{sec:pos_sen} does not show significant spatial dependence for post-fabrication error sensitivity.

\section{BlockFFTNet}
\label{app:hybrid}

\begin{figure}[h!]
    \centering
    \subfloat[]{%
    \label{fig:block_fft_arch}%
    \includegraphics[height=1.8in]{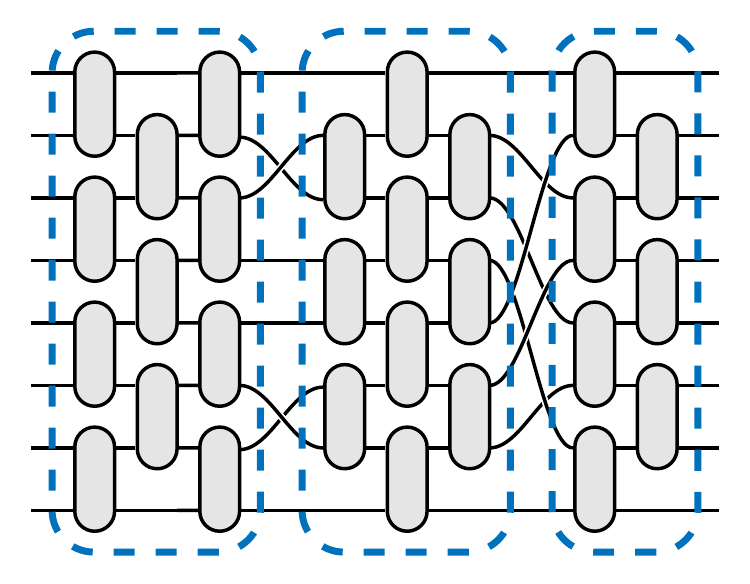}}%
    \subfloat[]{%
    \label{fig:block_fft_phase}%
    \includegraphics[height=1.8in]{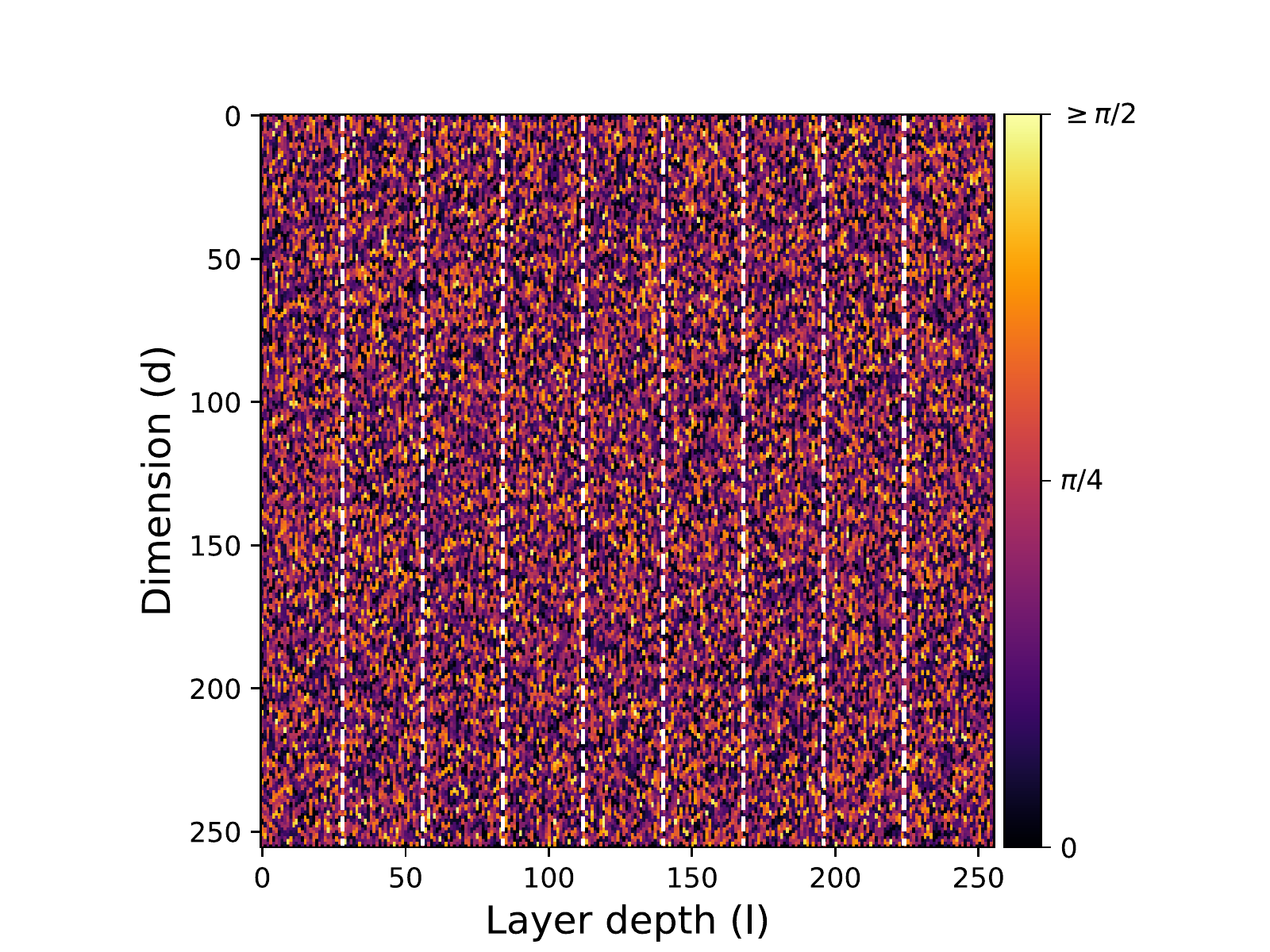}}%
    \caption{a) A schematic of BlockFFTUnitary. Blocks of MZIs in dashed, blue boxes are similar to GridNet. The crossing waveguide, similar to those in FFTNet are between the blocks. b) The distribution of phases after being trained. The dashed white lines denote the locations of the crossing waveguides.}
\end{figure}

\begin{figure}
    \centering
    \includegraphics[width=.8\columnwidth]{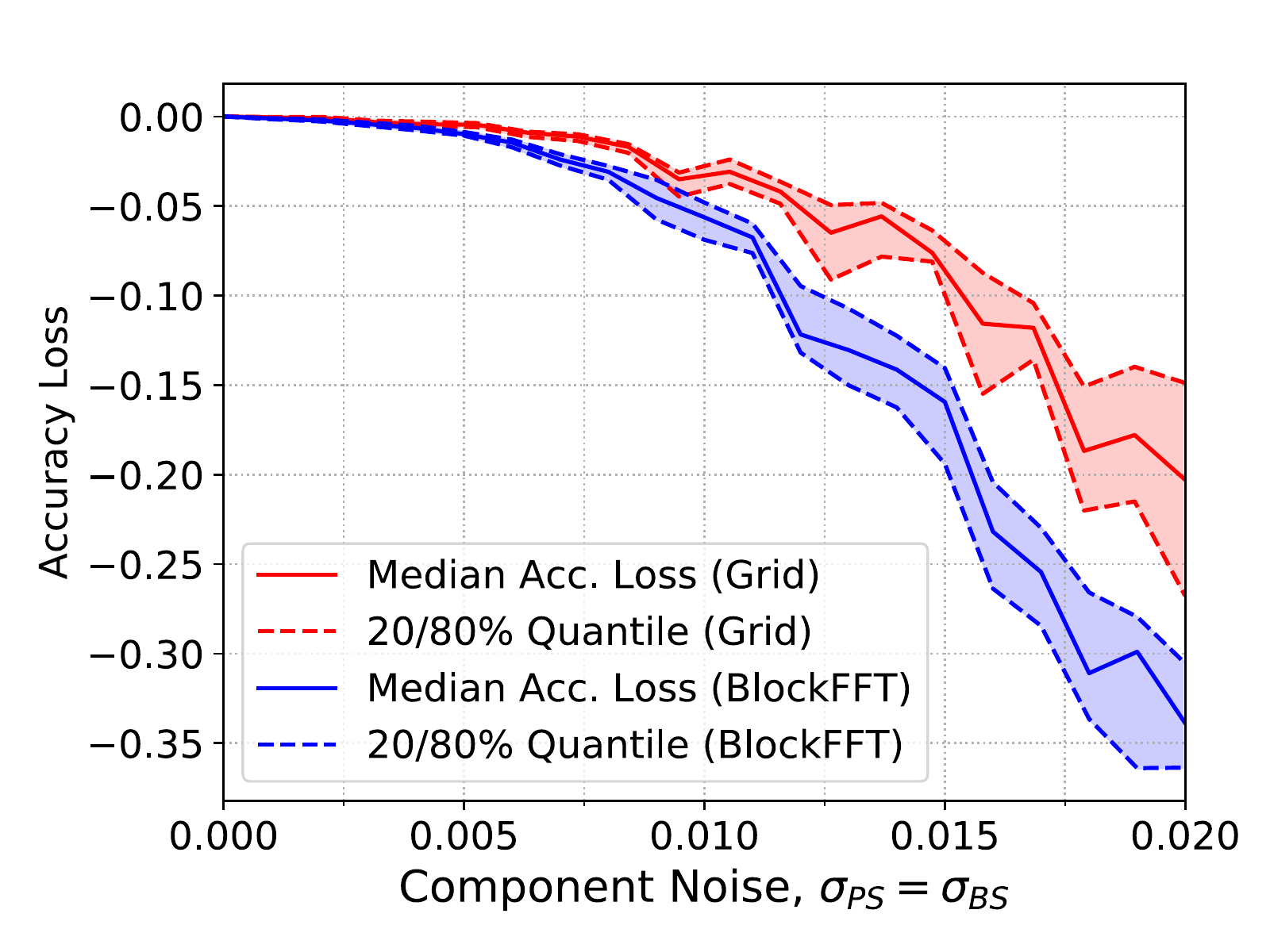}
    \caption{No improvement in robustness to imprecision is seen with BlockFFTNet over GridNet. In fact, there is a significant decrease.}
    \label{fig:blockfft_compare}
\end{figure}

We introduce a network with similar depth as GridUnitary but with non-local, crossing waveguides in between as those seen in FFTUnitary (Fig. \ref{fig:block_fft_arch}). This is similar to the coarse-grained rectangular design mesh in \cite{pai2018matrix} which was motivated to produce a spatially uniform distribution of phase and thus better tolerance for post-fabrication optimization. We also empirically observe that when incorporated as part of a ONN (BlockFFTNet), the distribution of phases are also uniformly distributed (Fig. \ref{fig:block_fft_phase}).
We directly demonstrate that better tolerance for post-fabrication optimization does not directly to better error-resistance for a network optimized pre-fabrication. The accuracy loss due to increasing imprecision is shown in Fig. \ref{fig:blockfft_compare}.

\section{FFT algorithm and convolution}
\label{app:FFT}
We show that the actual Cooley-Tukey FFT algorithm can be implemented with appropriate configurations of the phases of FFTUnitary multiplier.

If we denote the input as $x_n \in \mathbb C^N$, its Fourier transform is 
\begin{align}
X_k = \frac1{\sqrt{N}} \sum_{m=0}^{N-1} x_n e^{-\frac{2\pi i}{N} n k}.
\end{align}
The FFT algorithm, in short, is to rewrite the above as
\begin{align}
X_k &= \frac1{\sqrt{2}} \left(E_k + e^{- \frac{2\pi i}{N} k} O_k\right)\\
X_{k+N/2} &= \frac1{\sqrt{2}}\left(E_k - e^{- \frac{2\pi i}{N} k} O_k\right).
\end{align}
Here, we have defined $O_k$ and $E_k$ to be the Fourier transform on the odd and even elements of $x_n$ respectively. The calculation of $E_k$ and $O_k$ are done recursively. For $N = 2^K$, a total of $K$ iterations are needed. It is well known that if $x_n$ is in bit-reversed order, the calculations can be done in place.

Furthermore, in matrix form,
$$
\begin{pmatrix}
X_k \\
X_{k+N/2}
\end{pmatrix}
=
\frac1{\sqrt{2}}
\begin{pmatrix}
1 & e^{- \frac{2\pi i}{N} k}\\
1 & -e^{- \frac{2\pi i}{N} k}\\
\end{pmatrix}
\begin{pmatrix}
E_k\\
O_k
\end{pmatrix} \equiv
U_k
\begin{pmatrix}
E_k\\
O_k
\end{pmatrix}.
$$
 
From Eq. \eqref{eq:U_MZI}, we note that $U_k = U_{MZ}(\theta = \pi/2, \phi = 2\pi k /N)$, up to some global phase. Therefore, if $x_n$ is in bit-reversed order, and passed through a FFTUnitary multiplier where the $k$th layer is configured with $\theta = \pi/2, \phi = 2\pi k/ N$, FFT can be performed. 

Going further, a convolution can be easily performed through multiplication of the Fourier transformed signal by the Fourier transformed convolutional kernel, followed by a inverse Fourier transform.

\end{appendices}
\end{document}